\documentclass{article}

\usepackage[preprint]{neurips_2026}

\usepackage[utf8]{inputenc} 
\usepackage[T1]{fontenc}    
\usepackage{hyperref}       
\usepackage{url}            
\usepackage{booktabs}       
\usepackage{amsfonts}       
\usepackage{nicefrac}       
\usepackage{microtype}      
\usepackage{subcaption}
\usepackage{amsmath}
\usepackage{tikz}
\usetikzlibrary{positioning, fit, shapes.geometric, arrows.meta, calc, shadows}
\usepackage{algorithm}
\usepackage{algorithmic}
\usepackage{float}
\usepackage{amssymb}
\usepackage{amsthm}
\usepackage{siunitx}
\usepackage{booktabs}
\usepackage{mathtools}
\usepackage{multirow}
\usepackage{makecell}
\usepackage[dvipsnames]{xcolor} 
\usepackage{tcolorbox}
\tcbuselibrary{theorems}
\newcommand{\valpm}[2]{\ensuremath{#1} \scriptsize $\pm #2$}
\newcommand{\valpmbf}[2]{\ensuremath{\mathbf{#1}} \scriptsize $\pm #2$}
\newcommand{\valpmul}[2]{{\ensuremath{\underline{#1}} \scriptsize $\pm #2$}}

\usepackage{xspace}
\newcommand{\trueop}{\ensuremath{\mathcal{G}^\dagger}\xspace}
\newcommand{\paramnop}{\ensuremath{\mathcal{G}_\theta}\xspace}
\newcommand{\inu}{\ensuremath{\mathbf{u}}\xspace}
\newcommand{\ingeom}{\ensuremath{\mathbf{a}}\xspace}
\newcommand{\outv}{\ensuremath{\mathbf{v}}\xspace}
\newcommand{\residual}{\ensuremath{\boldsymbol{\delta}}\xspace}
\newcommand{\xb}{\mathbf{x}}
\newcommand{\augz}{\ensuremath{\mathbf{\bar{z}}}\xspace}
\newcommand{\augZ}{\ensuremath{\mathbf{\bar{Z}}}\xspace}
\newcommand{\phib}{\ensuremath{\boldsymbol{\phi}}\xspace}
\newcommand{\psib}{\ensuremath{\boldsymbol{\psi}}\xspace}

\hypersetup{hidelinks}

\title{Geometry-Aware Post-Hoc Uncertainty Quantification in Operator Learning}

\author{
Oriol Vendrell-Gallart \quad Nima Negarandeh \quad Ramin Bostanabad\\
Department of Mechanical and Aerospace Engineering\\
University of California, Irvine\\
Irvine, CA 92617 \\
\texttt{\{ovendrel,nnegaran,raminb\}@uci.edu} \\
}

\begin{document}

\maketitle

\begin{abstract}

Neural operators provide fast surrogates for PDEs but their deterministic predictions limit their use in tasks requiring uncertainty quantification (UQ), especially under geometric variability. Existing approaches primarily model uncertainty in network parameters, largely overlooking the geometry-aware representations learned by the operator itself.
We propose REEF-GP (Residual on Embedded Features Gaussian Process), a post-hoc UQ framework that fits a GP to the residuals of a frozen neural operator whose internal embeddings define the kernel feature space. Rather than learning a separate feature map, REEF-GP adapts the operator’s intrinsic coordinate-feature representations to construct geometry-aware uncertainties.
To ensure stability and scalability on unstructured domains, REEF-GP incorporates spectral-normalized projections, heteroscedastic geometry-aware noise, and efficient subset-based training that avoids restrictive low-rank approximations.
Across five PDE benchmarks with varying geometries, REEF-GP preserves predictive accuracy while achieving calibrated uncertainty estimates competitive with deep ensembles but at a fraction of their cost. Our approach remains robust under geometric distribution shift, with uncertainty concentrating in physically meaningful regions (e.g., shock fronts).
Our results demonstrate that accurate and scalable post-hoc UQ for neural operators can be achieved directly in their learned feature space, offering a practical alternative to parameter-centric approaches.

\end{abstract}

\section{Introduction}
Neural operators are increasingly used to approximate the solution of partial differential equations (PDEs) in domains such as climatology \cite{pathak2022fourcastnetglobaldatadrivenhighresolution, 10.5555/3618408.3618525}, fluid dynamics \cite{li2021fourierneuraloperatorparametric, rahman2024pretraining}, plasma physics \cite{gopakumar2023fourierneuraloperatorplasma, carey2025neuraloperatorsurrogatemodels}, and solid mechanics \cite{khorrami2025physicsencodedfourierneuraloperator, jeyaraj2025neuraloperatorbasedhybrid, 10.5555/3648699.3649087}. Driven by this growing adoption, recent years have seen substantial advances in their architectural design and training strategies to bridge the accuracy gap with traditional solvers while improving scalability and accommodating PDE solutions over varying geometries. In this regard, Transolver and its extensions \cite{wu2024transolverfasttransformersolver, luo2025transolveraccurateneuralsolver, zhou2026transolver3scalingtransformersolvers} stand out as prominent examples that leverage transformer-based backbones to achieve high accuracy, scale effectively to large problems, and natively operate on point clouds or mesh-based datasets.

Despite recent successes, most neural operators are built deterministically and provide point estimates that overlook uncertainties. This limitation hinders their adoption in scientific applications where UQ is critical \cite{mouli2024usinguncertaintyquantificationcharacterize, PSAROS2023111902} or in downstream tasks that rely on probabilistic predictions (e.g., adaptive data collection \cite{musekamp2025active}). Existing methods for addressing this issue broadly fall into train-time and post-hoc categories. Train-time approaches include the gold-standard deep ensembles \cite{10.5555/3295222.3295387}, where predictions of multiple independent models are combined together; Bayesian approximations via stochastic regularization techniques such as MC Dropout \cite{10.5555/3045390.3045502}; and inherently probabilistic models like DINOZAUR \cite{matveev2025lightweightdiffusionmultiplieruncertainty}, which uses a diffusion multiplier to introduce stochasticity to Fourier neural operators (FNOs) \cite{li2021fourierneuraloperatorparametric}.

\begin{figure}[!t]
    \centering
        \includegraphics[width=\linewidth, trim=7mm 0mm 6mm 0mm, clip]{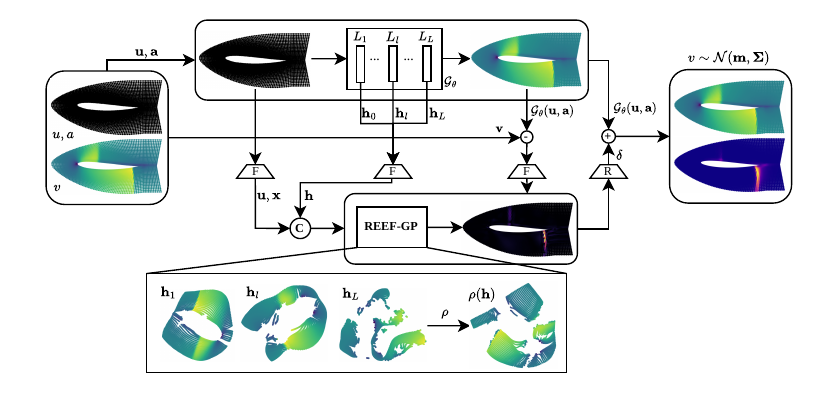}
    \caption{\textbf{REEF-GP architecture.} \paramnop is a frozen neural operator that maps the input functions $u,a$ to the output function $v$. REEF-GP models \paramnop's residual discrepancy by operating on the concatenation (C) of $\inu, \mathbf{x}, \mathbf{h}$ in the functional regression form (F). Internally, it transforms ($\rho$) the learned geometry-aware embeddings of \paramnop's internal layers $(\mathbf{h_1}, \mathbf{h_l}, \mathbf{h_L})$ to a new space where the kernel operates. In this example, the embedding in the feature-coordinate kernel space tears off the geometry exactly at the shock wave locations. The discrepancy $\delta$ is added to the base prediction after recovering (R) the point cloud form to obtain the output distribution.}
    \label{fig:flowchart}
\end{figure}

In contrast to train-time UQ, post-hoc methods add probabilistic components to a pretrained model at inference time. This alternative is particularly attractive since training neural operators can be an expensive and resource-intensive process, especially for inherently probabilistic variants. Last-layer Laplace approximation \cite{NEURIPS2021_a7c95857} is a popular and generic post-hoc technique that has been recently applied to operator learning \cite{magnani2025linearization}. Our method belongs to this post-hoc category but, unlike Laplace approximation, it builds Gaussian processes (GPs) \cite{10.7551/mitpress/3206.001.0001} whose mean and kernel functions are constructed around a pretrained neural operator, see Figure \ref{fig:flowchart}.

Kernel methods and GPs \cite{MORA2025117581,RN1881,lowery2024kernel} have recently been used for operator learning and benefit from theoretical guarantees and connections to Bayesian inference \cite{batlle2025error}. However, they suffer from scalability issues related to dataset size and dimensionality, especially in the case of mesh-based data. Additionally, Bayesian inference in infinite-dimensional spaces is inherently \textit{brittle} \cite{doi:10.1137/130938633}: unlike in finite settings, if the assumed prior is even slightly misspecified regarding the operator's regularity, the posterior may converge to an incorrect solution with high confidence rather than contracting to the truth. Consequently, standard approximations unaware of the underlying infinite dimensional nature of the problem introduce inductive biases that are liable to yield uncalibrated predictions. Reliable UQ therefore requires not just approximating the posterior of a fixed model, but actively learning from data the prior that best describes the geometry of the operator \cite{OWHADI201922}.

We introduce REEF-GP, a novel post-hoc UQ framework that models the discrepancy of a frozen pretrained neural operator with respect to the true solution as a function of both spatial coordinates and latent geometric features. Our main contributions are as follows:
\begin{itemize}
    \item We design a geometry-aware deep kernel that leverages the internal representations of a pretrained operator. These hidden layers encode deformed representations of the input geometry and REEF-GP reuses them rather than imposing a prior of its own.
    \item We develop training and inference procedures based on stochastic subset optimization and product of experts to make REEF-GP practical at the scale of operator learning datasets.
    \item We demonstrate competitive calibration on five challenging 2D and 3D benchmarks, including settings with geometric distribution shift.
\end{itemize}

\section{Background and Related Work}

\subsection{Geometry-Aware Operator Learning}

Let $\mathcal{U}$ and $\mathcal{V}$ be two Banach spaces of functions defined on a reference bounded domain $\Omega \subset \mathbb{R}^d$, where $d$ denotes the spatial dimension. Each PDE instance is associated with a continuous geometric descriptor $a \in \mathcal{A}$, which corresponds to a specific physical domain $D_a \subseteq \Omega$. For any input function and geometry pair $(u,a) \in \mathcal{U} \times \mathcal{A}$, the corresponding PDE solution $v \in \mathcal{V}$ is defined on $D_a$ and satisfies the governing equations:
\begin{equation}
\begin{aligned}
    \mathcal{P}(v, u)(\xb) &= 0, \quad \xb \in D_a, \\
    \mathcal{B}(v, u)(\xb) &= 0, \quad \xb \in \partial D_a,
\end{aligned}
\end{equation}
where $\mathcal{P}$ and $\mathcal{B}$ denote the differential and boundary operators, respectively.
Assuming a unique solution operator \trueop exists, we can write:
\begin{equation}
    \trueop: \mathcal{U} \times \mathcal{A} \to \mathcal{V}.
\end{equation}


\paragraph{Neural Operators.}
Under this framework, the objective is to learn the infinite-dimensional operator \trueop given some data. To this end, these models construct a parametric operator \paramnop that approximates \trueop by minimizing the empirical risk over the dataset $\mathcal{D}$ containing $M$ triplets $\{(u_i,a_i,v_i)\}_{i=1}^M$, where $v_i = \trueop(u_i, a_i)$. While \trueop acts on continuous spaces, in practice we only have access to discrete numerical evaluations $\{(\inu_i,\ingeom_i,\outv_i)\}_{i=1}^M$. For a given instance, the continuous domain is discretized as a mesh or point cloud $\ingeom_i = \{\mathbf{x}_j\}_{j=1}^N \subset D_{a_i}$ and $\outv_i = \{v_i(\xb_{i,j})\}_{j=1}^N \in \mathbb{R}^N$ collects the corresponding nodal solution values. 

A large body of work in operator learning has focused on designing architectures that can represent solution operators across discretizations and geometries. Early spectral approaches such as FNOs are particularly effective on regular discretizations \cite{li2021fourierneuraloperatorparametric} but more recent geometry-aware variants accommodate irregular domains, meshes, and point clouds \cite{10.5555/3648699.3649087,li2023geometryinformed,wu2024transolverfasttransformersolver}. Transformer-based architectures such as Transolver are especially relevant in this setting as they natively operate on general geometries and preserve spatially aligned hidden states that can later be exploited for UQ.

\subsection{Gaussian Processes}\label{sec:gps-for-operator-learning}

GPs \cite{10.7551/mitpress/3206.001.0001} provide a principled Bayesian framework for regression, yielding closed-form posteriors with calibrated uncertainty estimates. We apply them to operator learning via a \emph{functional regression} perspective \cite{MORA2025117581}: instead of directly learning \trueop, which outputs an infinite-dimensional function, we learn its evaluation functional, so the pointwise evaluation of the operator is represented by a finite-dimensional surrogate
\begin{equation}
\label{eq:gp_for_ol}
    \widetilde{\mathcal{G}}^\dagger: \mathbb{R}^{d_u} \times \mathbb{R}^{d_a} \times \Omega \to \mathbb{R}, \qquad (\inu, \ingeom, \mathbf{x}) \mapsto v(\mathbf{x}),
\end{equation}
where $\inu \in \mathbb{R}^{d_u}$ and $\ingeom \in \mathbb{R}^{d_a}$ are discrete encodings of the input function $u$ and geometry $a$, and $\mathbf{x} \in D_a$ is a query coordinate. Operator learning thus reduces to scalar regression over an augmented input $\mathbf{z} = (\inu, \ingeom, \mathbf{x})$. The complete derivation is provided in Appendix~\ref{appendix:gps-for-operator-learning}. Given $M$ samples each evaluated at $N$ spatial points, the training set contains $MN$ pairs $\{(\mathbf{z}_i, v_i)\}_{i=1}^{MN} = \{\mathbf{Z}, \mathbf{v}\}$, where each $\mathbf{z}_i = (\inu_{s(i)}, \ingeom_{s(i)}, \mathbf{x}_i)$ and $s(i)$ identifies the sample to which point $i$ belongs. A GP prior over the latent function $f(\mathbf{z})$ takes the form
\begin{equation}
    f(\mathbf{z}) \sim \mathcal{GP}\big(m(\mathbf{z}; \boldsymbol{\beta}),\ k(\mathbf{z},\mathbf{z}'; \phib)\big),
\end{equation}
with parametric mean and covariance functions $m(\cdot)$ and $k(\cdot,\cdot)$. Given noisy observations $v_i = f(\mathbf{z}_i) + \epsilon$ with $\epsilon \sim \mathcal{N}(0,\lambda^2)$, the hyperparameters $\{\boldsymbol{\beta}, \phib, \lambda^2\}$ are typically optimized via maximum likelihood estimation. The posterior mean and covariance at a test input $\mathbf{z}_*$ then admit the closed forms
$\bar{m}(\mathbf{z}_*) = m(\mathbf{z}_*) + k(\mathbf{z}_*,\mathbf{Z})(\mathbf{K}+\lambda^2 \mathbf{I})^{-1}(\mathbf{v}-m(\mathbf{Z}))$
and
$\bar{k}(\mathbf{z}_*,\mathbf{z}_*') = k(\mathbf{z}_*,\mathbf{z}_*') - k(\mathbf{z}_*,\mathbf{Z})(\mathbf{K}+\lambda^2 \mathbf{I})^{-1}k(\mathbf{Z},\mathbf{z}_*')$.

\paragraph{Scalability and Kernel Design.}
Standard GPs are fundamentally limited by their $\mathcal{O}(N^3)$ time and $\mathcal{O}(N^2)$ memory complexity, making exact inference intractable for operator learning where datasets can easily contain millions of discretization points. Moreover, commonly used scalable GP approximations can introduce structural assumptions that are poorly matched to geometry-varying PDE surrogates. For instance, tensor-product kernels \cite{10.5555/3045118.3045307, pmlr-v89-zhe19a} are naturally tied to regular grids and impose separable covariance structure, which can be restrictive for non-stationary and anisotropic solution fields on irregular domains. Likewise, inducing-point approximations such as Nyström \cite{NIPS2000_19de10ad}, SGPR \cite{pmlr-v5-titsias09a}, and SVGP \cite{pmlr-v38-hensman15} rely on low-rank approximations that can limit the GP's flexibility.

Prior GP-based approaches to operator learning involve fixed-geometry settings \cite{RN1881, MORA2025117581} where the input dimensionality remains moderate and Kronecker structure can be exploited for scalability. These advantages disappear with geometric variability. A natural alternative is deep kernel learning (DKL) which, when combined with scalable approximations, has enabled GPs to address large-scale and high-dimensional problems. However, in settings governed by strong structural constraints (e.g., PDE-driven  problems) kernel design is critical for achieving both expressivity and well-calibrated uncertainty \cite{pmlr-v51-wilson16}. In this work, we address this challenge by leveraging the learnt embeddings of a pretrained neural operator, thus encoding geometry and physics-aware structure directly into the GP.

\subsection{Scalable Uncertainty Quantification for Neural Operators}

From a Bayesian perspective, UQ in deep networks such as neural operators amounts to characterizing the predictive distribution induced by uncertain model parameters or representations. In principle, this requires marginalizing over a posterior involving millions of parameters. Since this is computationally intractable, practical UQ relies on scalable approximations that are either train-time or post-hoc.

\paragraph{Train-time UQ.}
These methods consider uncertainties during training, typically via weight-space stochasticity or repeated retraining as in deep ensembles \cite{10.5555/3295222.3295387}. Computationally efficient alternatives include SWAG \cite{10.5555/3454287.3455466}, which performs approximate Bayesian model averaging using weight trajectories collected during training, and Monte Carlo (MC) dropout \cite{10.5555/3045390.3045502}, where uncertainty is estimated via repeated stochastic forward passes with dropout activated at inference time. These ideas have also been adapted to operator learning; for example, Probabilistic Neural Operator (PNO) \cite{lte2024probabilistic, blte2025probabilistic} uses dropout during training to generate functional samples and optimize an energy-score objective. In broader deep-learning settings, related architectural modifications such as Spectral-normalized GPs \cite{JMLR:v24:22-0479} have also been proposed to improve uncertainty estimation.

\paragraph{Post-hoc UQ.}
To preserve the computational efficiency of deterministic training, post-hoc methods augment a frozen pretrained base model for UQ. The most widely used approach is the Laplace approximation \cite{10.1162/neco.1992.4.3.415, ritter2018a, NEURIPS2021_a7c95857}, which fits a local Gaussian posterior around the maximum a posteriori (MAP) estimate of the weights (for scalability, this strategy is often restricted to the last layer). Laplace approximation has also been explored for neural operators \cite{DBLP:journals/corr/abs-2208-01565, weber2024uncertainty}, and more recently LUNO \cite{magnani2025linearization} applies it to the internal spectral representations and propagates them through a linearized upstream network. Beyond operator learning, Deep Vecchia Ensembles (DVE) \cite{pmlr-v258-jimenez25a} move from weight-space to feature-space by fitting a GP on internal neural representations.

While train-time methods benefit from rich epistemic exploration, they are prohibitively expensive for large-scale operator learning on unstructured meshes. In contrast, post-hoc UQ methods are cheaper but beyond DVE they mostly rely on local weight-space approximations instead of fully exploiting the geometry-aware structure encoded in the pretrained operator. These limitations motivate uncertainty models that operate directly in the coordinate and feature space of frozen neural operators.

\section{Method: REEF-GP}\label{sec:method}

Most existing UQ methods for neural operators act in weight space, either during training or through post-hoc local approximations. With REEF-GP (Figure~\ref{fig:flowchart}) we pursue a complementary perspective: rather than perturbing the model parameters, we model the structural inadequacy of a pretrained neural operator directly in the input-coordinate space induced by the operator and its internal representations.
More specifically, we approach the problem from the perspective of the Kennedy--O'Hagan (KOH) discrepancy framework \cite{10.1111/1467-9868.00294}. While KOH was originally introduced for Bayesian calibration of computer models and fusing high- and low-fidelity datasets \cite{RN270,RN334}, we reformulate it for post-hoc UQ in operator learning. We treat the pretrained neural operator \paramnop as a deterministic computer model whose parameters remain strictly frozen. Since \paramnop is an approximation to \trueop, there exists a residual discrepancy between the two which we interpret as \emph{model inadequacy}.

Building on the functional regression view introduced in Section~\ref{sec:gps-for-operator-learning}, we write the true solution evaluated at a spatial query coordinate $\xb \in D_a$ as:
\begin{equation}
    v(\xb) = \mathcal{G}_\theta(\inu,\ingeom)(\xb) + \delta(\inu,\ingeom,\xb) + \epsilon,
\end{equation}
where $\delta(\cdot)$ denotes the model inadequacy and $\epsilon$ captures stochastic variability or observational noise.

Unlike classical KOH formulations which treat the computer model as a black box, we exploit its internal structure. A neural operator can be written as a composition of $L$ latent transformations:
\begin{equation}
    \paramnop(\inu,\ingeom) = \mathcal{L}_{L} \circ \mathcal{L}_{L-1} \circ \dots \circ \mathcal{L}_1(\inu,\ingeom).
\end{equation}
Therefore, $\delta(\cdot)$ is the result of cumulative approximation errors propagated through the network's internal states (this observation is consistent with the representational limits and spectral bias in deep learning \cite{pmlr-v97-rahaman19a}).
Additionally, recent analyses indicate that compared to the highly constrained output layer, intermediate representations often contain strictly richer and less-compressed information regarding the underlying physics \cite{skean2025layer, pmlr-v258-jimenez25a}. In the case of operator learning, we demonstrate this \textit{multiscale} representation with a detailed layer-wise analysis in Appendix \ref{appendix:kernel_flows}, where we also illustrate the layer-wise geometry deformation using t-SNE \cite{JMLR:v9:vandermaaten08a}.


Motivated by this multiscale error structure, we enrich the discrepancy model with hidden representations extracted from the frozen \paramnop. Let $\mathbf{h}_l(\inu,\ingeom,\xb)$ denote the spatially aligned hidden feature at layer $l$ and query coordinate $\xb$. We select $L' \leq L$ informative layers (the selection procedure is described in Appendix~\ref{appendix:layer_selection} with ablations in Appendix~\ref{appendix:ablation:layer_selection}) and concatenate them into the augmented state 
$\mathbf{h}(\inu,\ingeom,\mathbf{x}) = \bigoplus_{l=1}^{L'} \mathbf{h}_l(\inu,\ingeom,\mathbf{x})$. 
We then refine the discrepancy model by replacing the black-box correction term with a feature-aware alternative defined on the coordinate-feature state: 
\begin{equation}
    v(\xb) = \mathcal{G}_\theta(\inu,\ingeom)(\xb) + \delta(\inu, \xb, \mathbf{h}(\inu,\ingeom,\xb)) + \epsilon.
\end{equation}
This yields a discrepancy model conditioned on location and the local regime given by the operator.

\subsection{Probabilistic Discrepancy Modeling}

To obtain calibrated uncertainties, we model the feature-aware discrepancy probabilistically by first decomposing it into two parts: a latent noiseless residual correction function $f(\cdot)$ and a heteroscedastic noise term $\varepsilon(\cdot)$ that accounts for the unexplained variability:
\begin{equation} \label{eq delta terms}
    \delta(\inu,\xb, \mathbf{h}(\inu,\ingeom,\xb))
    =
    f(\inu, \xb, \mathbf{h}(\inu,\ingeom,\xb))
    +
    \varepsilon(\inu, \xb,\mathbf{h}(\inu,\ingeom,\xb)).
\end{equation}
Then, we place a zero-mean GP prior over the residual, that is $f \sim \mathcal{GP}(0, k_{\phib}(\augz, \augz'))$ where $\augz$ is the augmented state $\augz = (\inu, \xb, \mathbf{h}(\mathbf{u},\mathbf{a}, \xb))$ and $k_{\phib}$ is the kernel with hyperparameters $\phib$. 
Next, we jointly model the unresolved error and the independent observational noise as:
\begin{equation} \label{eq error terms}
    \varepsilon(\augz) + \epsilon
    \sim
    \mathcal{N}\!\left(0,\sigma_{\psib}^2(\augz) + \sigma_n^2\right),
\end{equation}
where $\sigma_n^2$ is a global observation-noise term modeled as $\epsilon \sim \mathcal{N}(0, \sigma_n^2)$. Although many PDE benchmarks are effectively deterministic, retaining $\sigma_n^2$ allows our framework to accommodate stochastic simulators or measurement noise in experimental settings. 
Moreover, we parametrize $\sigma_{\psib}^2$ by a spectral-normalized \cite{JMLR:v24:22-0479} dense network that maps $\augz$ to a strictly positive value. The spectral normalization mitigates feature collapse which, in turn, enables a GP to surrogate $\delta(\cdot)$. 

With the independent Gaussian priors and Equations \ref{eq delta terms} and \ref{eq error terms}, the marginal prior for the solution value at a query point $\xb$ is:
\begin{equation}
    v(\xb) \sim \mathcal{N}\Big(
        \mathcal{G}_\theta(\mathbf{u},\mathbf{a})(\xb),\;
        k_{\phib}(\augz,\augz) + \sigma_{\psib}^2(\augz) + \sigma_n^2
    \Big).
\end{equation}
The kernel design, hyperparameter estimation, and inference are detailed in the next sections. 

\subsection{Kernel Design for Infinite-dimensional Inference}\label{sec:kernel}

The rationale behind defining the discrepancy $\delta$ as a function of the operator's internal state is rooted in a fundamental theoretical challenge: the inherent brittleness of Bayesian inference in infinite-dimensional spaces \cite{doi:10.1137/130938633}. In finite-dimensional settings, posterior contraction is often robust to moderate prior misspecification, but in function spaces even mild mismatch between the prior regularity and the target operator can lead to highly confident yet incorrect inference.

To avoid imposing a misspecified prior, we draw inspiration from Kernel Flows (KF) \cite{OWHADI201922} which advocates for learning the kernel geometry from data without the need of relying on physics informed priors \cite{JMLR:v24:22-0676}. Our key modification is to avoid learning the manifold from scratch: we assume that the pretrained neural operator has already deformed the original input-geometry space into a physics-aware latent representation through its hidden layers. In Appendix~\ref{appendix:kernel_flows} we provide a more formal motivation for this latent geometry and its connection to KF and DKL.

We therefore construct a deep kernel \cite{pmlr-v51-wilson16} over the augmented state (see Appendix \ref{appendix:kernel_construction} for architecture details). To respect the distinct topological roles of spatial location, input constraints, and latent operator state, we leverage kernel closure properties and define the covariance as the product kernel:
\begin{equation}
    k_{\phib}(\augz,\augz')
    =
    k_{\text{space}}(\xb,\xb')
    \cdot
    k_{\text{fun}}(\inu,\inu')
    \cdot
    k_{\text{latent}}(\mathbf{h},\mathbf{h}').
\end{equation}
Here, $k_{\text{space}}$ acts on physical coordinates, $k_{\text{fun}}$ acts on the input function representation when present, and $k_{\text{latent}}$ acts on the hidden operator embeddings. 
For the latent component, we further define:
\begin{equation}
    k_{\text{latent}}(\mathbf{h},\mathbf{h}')
    =
    k_{\text{base}}\!\left(\rho(\mathbf{h}),\rho(\mathbf{h}')\right),
\end{equation}
where $k_{\text{base}}$ is a stationary kernel and $\rho(\cdot)$ is a warping function parameterized with a spectral-normalized MLP. This construction yields a non-stationary prior in the original coordinate space: spatial correlations are modulated by similarity in the learned latent state, allowing the discrepancy model to adapt to different local physical regimes while remaining anchored to the pretrained operator's internal geometry.

\subsection{Scalable Training}

Under the functional regression formulation of Section~\ref{sec:gps-for-operator-learning}, the training set contains \(MN\) elements. In operator learning $MN$ is often in the order of millions and so exact GP training is infeasible due to \(\mathcal{O}((MN)^3)\) time and \(\mathcal{O}((MN)^2)\) memory requirements. Hence, again inspired by KF \cite{OWHADI201922}, we develop a stochastic subset strategy that jointly optimizes all model hyperparameters. 
At each optimization step, we randomly sample a subset of \(N_s \ll MN\) points and evaluate \residual defined as the \(N_s\)-dimensional vector of structural residuals with entries
$\delta_i = v(\xb_i) - \mathcal{G}_\theta(\inu_{s(i)},\ingeom_{s(i)})(\xb_i)$.
Then, we update the hyperparameters using mini-batch negative log marginal likelihood as the loss:
\begin{equation}
    \mathcal{L}(\phib,\psib)
    =
    \frac{1}{2}\residual^\top \mathbf{K}_{\phib\psib}^{-1}\residual
    +
    \frac{1}{2}\log|\mathbf{K}_{\phib\psib}|
    +
    \frac{N_s}{2}\log(2\pi),
\end{equation}
where $\mathbf{K}_{\phib\psib} = k_{\phib}(\augZ,\augZ) + \mathrm{diag}\!\left(\sigma_{\psib}^2(\augZ)\right) + \sigma_n^2\mathbf{I}$ is the tractable $N_s \times N_s$ joint covariance matrix evaluated at the augmented states $\augZ$. By drawing random $N_s$-sized subsets at each epoch, we strictly decouple the memory constraints from the true mesh resolution $N$. In our studies, this training procedure combined with spectral normalization and heteroscedastic noise remains stable. The implementation, empirical validation and optimization settings are detailed in Appendix~\ref{appendix:training_algorithm}.

We highlight that the KF loss introduced in \cite{OWHADI201922} discards UQ and focuses on building kernel \textit{interpolants} whose accuracy is robust to the training data size. We borrow the notion of robustness behind KF loss and extend it to accommodate UQ. In Appendix~\ref{appendix:kernel_flows} we further elaborate in this regard and detail how REEF-GP differs from standard DKL with inducing points. 

\subsection{Inference via Generalized Product of Experts}

Conditioning the GP on $MN$ training points is infeasible for fast inference. To address this issue, we use generalized Product of Experts (gPoE) \cite{cao2015generalizedproductexpertsautomatic} where we first randomly draw $K$ support subsets of the data to construct $K$ independent GP experts that share the same hyperparameters. We then aggregate their predictive means and variances in precision space to obtain the posterior of our discrepancy surrogate. Finally, this posterior is added to the frozen base operator's prediction to form the final predictive distribution. This formulation strictly bounds the test-time memory footprint to $\mathcal{O}(N_s^2)$ per expert. Full mathematical details of the gPoE aggregation are provided in Appendix~\ref{appendix:inference_algorithm}.

\section{Experiments}\label{sec:experiments}

\paragraph{Base Neural Operator.} 

We use Transolver \cite{wu2024transolverfasttransformersolver} as the base deterministic neural operator in all experiments. Unlike operators with global spectral filters (e.g., FNOs), Transolver preserves pointwise spatial correspondence between blocks. This allows us to extract the spatially-aligned hidden features $\mathbf{h}(\inu, \ingeom, \xb)$ required for our augmented state. By keeping the base architecture fixed, we isolate the effects of our UQ formulation. Hyperparameter configurations are detailed in Appendix \ref{appendix:base_no}.

\paragraph{Benchmarks.} 

We use a combination of 2D (\emph{Elasticity, Airfoil}, and \emph{Pipe}) and 3D (\emph{ShapeNet Car} and \emph{Ahmed Body}) datasets involving point clouds. In the main text we present results on the representative 2D \emph{Airfoil} and 3D \emph{ShapeNet Car} benchmarks; full results across all five datasets are reported in Appendix~\ref{appendix:additional_results}. Dataset details are included in Appendix~\ref{appendix:benchmarks}.

\paragraph{Baselines.} 

We compare REEF-GP against six UQ baselines: (1) Deep Ensembles, (2) MC Dropout, (3) PNO, (4) Input Perturbation, (5) Laplace Approximation (LUNO-LA), and (6) Deep Vecchia Ensemble (DVE-spatial). All baselines use the same pretrained Transolver architecture. Implementation details are provided in Appendix~\ref{appendix:uq_baselines} and evaluation metrics are defined in Appendix~\ref{appendix:metrics}.

\begin{table}[!b]
\centering
\caption{\textbf{Evaluation metrics on representative 2D (Airfoil) and 3D (ShapeNet Car) benchmarks.} Lower is better ($\downarrow$). Best in \textbf{bold}, second best \underline{underlined}. Base is the deterministic Transolver without UQ; rankings exclude Base. REEF-GP is our approach. Metrics are defined in Appendix \ref{appendix:metrics}.}
\label{tab:results_main}
\setlength{\tabcolsep}{4pt}
\renewcommand{\arraystretch}{1.1}
\small
\begin{tabular}{l c c c c c}
\toprule
\textbf{Method} & \textbf{rL2} $\downarrow$ & \textbf{NLL} $\downarrow$ & \textbf{CRPS} $\downarrow$ & \textbf{NIS} $\downarrow$ & \textbf{ES} $\downarrow$ \\
\midrule
\emph{Airfoil (2D)} & (\%) &  & ($\times 10^{-4}$) & ($\times 10^{-2}$) &  \\
\midrule
Base (no UQ) & \valpm{1.24}{0.12} & -- & -- & -- & -- \\
\addlinespace[2pt]
Ensemble & $\mathbf{1.06}$ & $\underline{-3.46}$ & $\mathbf{28.63}$ & $\mathbf{3.89}$ & $\mathbf{0.35}$ \\
MC Dropout & \valpm{1.58}{0.21} & \valpm{-3.38}{0.15} & \valpm{44.88}{5.66} & \valpmul{5.55}{0.91} & \valpm{0.55}{0.08} \\
PNO & \valpm{2.32}{1.40} & \valpm{-3.28}{0.65} & \valpm{74.27}{56.12} & \valpm{8.26}{6.39} & \valpm{0.79}{0.47} \\
Perturbation & \valpm{3.33}{0.19} & \valpm{-3.32}{0.05} & \valpm{81.05}{5.67} & \valpm{9.71}{0.66} & \valpm{1.08}{0.06} \\
LUNO-LA & \valpmul{1.24}{0.12} & \valpm{-3.07}{0.82} & \valpmul{38.70}{2.50} & \valpm{6.19}{0.54} & \valpm{0.47}{0.04} \\
DVE-spatial & \valpm{2.09}{0.25} & \valpm{19.98}{7.32} & \valpm{88.41}{12.53} & \valpm{25.89}{3.09} & \valpm{0.92}{0.10} \\
\textbf{REEF-GP} & \valpmul{1.24}{0.12} & \valpmbf{-3.51}{0.06} & \valpm{39.80}{2.35} & \valpm{5.69}{0.47} & \valpmul{0.45}{0.05} \\
\midrule
\emph{ShapeNet Car (3D)} & (\%) &  &  &  &  \\
\midrule
Base (no UQ) & \valpm{8.62}{0.13} & -- & -- & -- & -- \\
\addlinespace[2pt]
Ensemble & $\mathbf{7.73}$ & $\underline{4.16}$ & $\mathbf{1.88}$ & $\underline{30.28}$ & $\mathbf{194.57}$ \\
MC Dropout & \valpm{9.04}{0.59} & \valpm{17.21}{4.21} & \valpm{2.52}{0.26} & \valpm{67.96}{9.04} & \valpm{267.68}{17.66} \\
PNO & \valpm{13.76}{1.74} & \valpm{4.98}{1.52} & \valpm{3.53}{0.50} & \valpm{51.10}{12.17} & \valpm{337.16}{43.04} \\
Perturbation & \valpm{8.62}{0.13} & \valpm{13.44}{1.14} & \valpm{2.28}{0.03} & \valpm{50.82}{1.10} & \valpm{224.73}{3.97} \\
LUNO-LA & \valpm{8.62}{0.13} & \valpm{33.55}{23.43} & \valpm{2.47}{0.13} & \valpm{73.47}{14.91} & \valpm{271.32}{12.10} \\
DVE-spatial & \valpm{13.08}{0.26} & \valpm{11.22}{3.03} & \valpm{3.76}{0.10} & \valpm{96.17}{10.43} & \valpm{396.70}{6.79} \\
\textbf{REEF-GP} & \valpmul{8.52}{0.14} & \valpmbf{2.84}{0.09} & \valpmul{2.05}{0.03} & \valpmbf{28.81}{1.65} & \valpmul{224.31}{4.18} \\
\bottomrule
\end{tabular}
\end{table}

\subsection{Results}

Table \ref{tab:results_main} reports the evaluation metrics on the two representative benchmarks. Results on the remaining three cases are deferred to Appendix \ref{appendix:additional_results} (Table \ref{tab:results_appendix}) and are consistent with the trends discussed here.

\paragraph{Predictive accuracy is preserved.} REEF-GP does not degrade the base operator's predictive accuracy in any of the five datasets (Tables \ref{tab:results_main}, \ref{tab:results_appendix}), and rL2 remains identical or very similar to the deterministic baseline. LUNO-LA also has this attractive property but Perturbation and all the train-time methods (MC Dropout, PNO, and DVE-spatial) incur substantial rL2 penalties relative to Base. The only method that (expectedly) improves rL2 over Base is Deep Ensembles but at the cost of substantial additional computation.

\paragraph{UQ matches Deep Ensembles on 3D benchmarks.}
On \emph{ShapeNet Car}, REEF-GP achieves NLL of $2.84$, outperforming even Deep Ensembles ($4.16$) and reducing NLL by an order of magnitude relative to the strongest post-hoc competitor LUNO-LA ($33.55$). 
On \emph{Ahmed Body}, our NLL is tied with the Deep Ensembles (Table~\ref{tab:results_appendix}). Regarding the rest of the metrics and datasets, REEF-GP is competitive to either the best or second-best performing baselines.

\begin{figure}[t] 
    \centering 
    
    \begin{subfigure}{\textwidth}
        \centering
        \includegraphics[width=\linewidth]{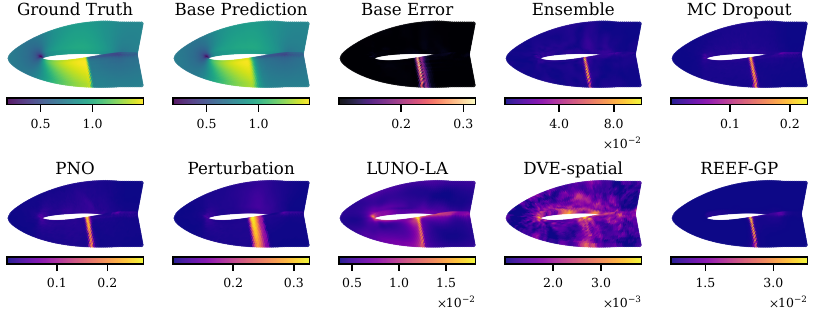} 
    \end{subfigure}
    
    \vspace{0.0em} 
    
    \begin{subfigure}{\textwidth}
        \centering
        \includegraphics[width=\linewidth]{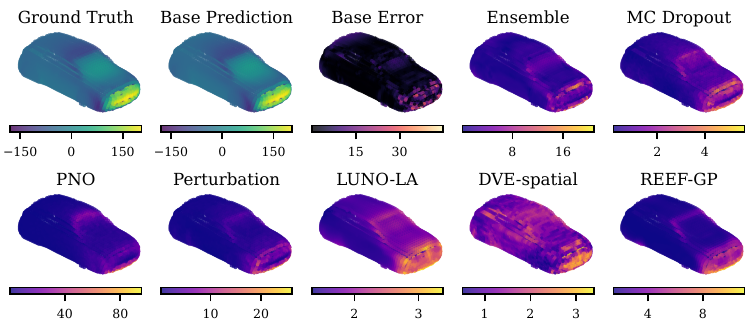} 
    \end{subfigure}
    
    \caption{\textbf{Predictive standard deviation fields on representative 2D (\emph{Airfoil}) and 3D (\emph{ShapeNet Car}) benchmarks.} Top two rows show an airfoil sample and bottom two rows a car sample. For each sample the rows show in order: ground truth, base prediction, base error and the standard deviation of a sample for each of the UQ baselines.}
    \label{fig:combined_samples}
\end{figure}

\textbf{Spatially coherent uncertainty maps.} 
Aggregate metrics can be inconclusive when best values shift across datasets and scoring rules. In geometric operator learning, spatial consistency between predicted uncertainty and prediction error is itself a meaningful signal. Figure~\ref{fig:combined_samples} shows per-method standard deviation fields on \emph{Airfoil} and \emph{ShapeNet Car} alongside the base operator's prediction and error. Our uncertainty concentrates around the shock fronts on \emph{Airfoil} and the front bumper region on \emph{Car}, closely tracking where the operator's error is largest. Calibration is therefore not just numerically tight but spatially meaningful.

\begin{figure}[!t]
    \centering
        \includegraphics[width=\textwidth]{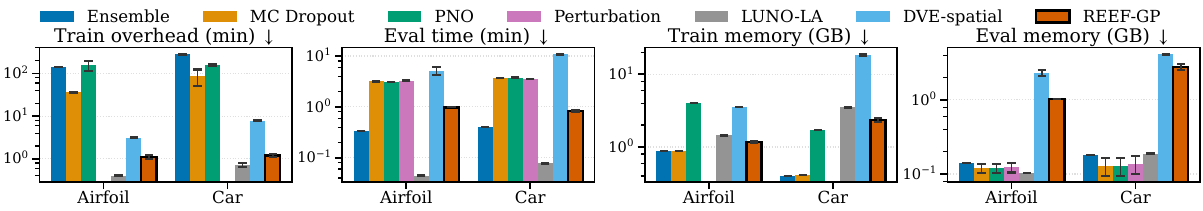}
    \caption{\textbf{Compute costs on representative 2D and 3D benchmarks:} 
    Train overhead, evaluation time, peak train memory, and peak evaluation memory on \emph{Airfoil} (2D) 
    and \emph{ShapeNet Car} (3D). All axes use log scale. REEF-GP's train 
    and evaluation times are competitive but its memory requirements are higher. We define training overhead as the additional wall-clock time required to add UQ capabilities to a pretrained deterministic base Transolver.}
    \label{fig:cost_panels_main_full}
\end{figure}

\textbf{Cost-quality tradeoff.} Figure~\ref{fig:cost_panels_main_full} compares compute cost across all UQ methods on \emph{Airfoil} and \emph{ShapeNet Car}. REEF-GP achieves training overhead an order of magnitude lower than Deep Ensembles and competitive with all post-hoc baselines. Eval time and train memory are competitive across methods, with REEF-GP sitting close to the median. The only tradeoff is in eval memory, where REEF-GP carries an $\mathcal{O}(N_s^2)$ cost from per-expert kernel matrices; absolute values nonetheless remain well within commodity GPU memory.

\subsection{Ablation Studies}

We conduct five ablations using the \emph{Airfoil} benchmark to dissect the impact of each architectural and methodological choice; full results, tables, and visualizations are deferred to Appendix~\ref{appendix:ablation_studies}. 
(i) \textbf{Layer selection} (Appendix~\ref{appendix:ablation:layer_selection}): our CKA-based three-layer selection $\{L_0, l^*, L_{\text{last}}\}$ matches concatenating the entire 8-layer Transolver stack at significantly lower memory cost, and substantially outperforms single-layer variants. 
(ii) \textbf{Size of training data} (Appendix~\ref{appendix:ablation:training_samples}): REEF-GP adapts gracefully to base operators of varying quality, with the predicted uncertainty correctly inflating to track the operator's deteriorating accuracy as $M_{\text{train}}$ decreases from $750$ to $50$, while the GP's $\text{rL2}$ closely follows the base operator's across two orders of magnitude in training-set size (Figure~\ref{fig:rl2_vs_samples}). 
(iii) \textbf{Stochastic subset size} (Appendix~\ref{appendix:ablation:subset_size}): performance is essentially insensitive to the per-expert support size $N_s$ across two orders of magnitude (from $500$ to $25{,}000$), with overlapping error bars on every probabilistic metric. We adopt $N_s = 5{,}000$ as a middle-of-the-range default that balances conditioning quality against the cubic Cholesky cost. 
(iv) \textbf{Number of gPoE experts} (Appendix~\ref{appendix:ablation:number_experts}): calibration is essentially insensitive to the expert count $K \in \{1, 5, 10\}$. 
(v) \textbf{Heteroscedastic vs. homoscedastic noise} (Appendix~\ref{appendix:ablation:noise}): both noise models yield comparable aggregate metrics, but the heteroscedastic variant produces sharper, error-localized uncertainty fields, motivating its use as our default.

\subsection{Geometric Out-Of-Distribution (OOD) Robustness}

To assess whether REEF-GP remains well-calibrated under geometric distribution shift, we partition each test set into four quartiles by maximum mean discrepancy (MMD) distance from the training distribution, ranging from Q1 (closest) to Q4 (most geometrically novel). A reliable UQ method should produce stable calibration across quartiles, inflating its predicted uncertainty as test geometries drift away from training. Across all benchmarks, REEF-GP provides the second-best rL2 (after Deep Ensemble) and its NLL negligibly changes from Q1 to Q4, while LUNO-LA's NLL and Perturbation's NLL significantly increase. Full per-quartile rL2 and NLL metrics, MMD construction details, and per-method comparisons are reported in Appendix~\ref{appendix:geometric_OOD} and Table~\ref{tab:results_quantile_ood}.

\section{Conclusions}\label{sec:conclusions}

Our results show that the coordinate–feature space induced by a pretrained neural operator can be used for post-hoc UQ. By fitting REEF-GP directly on the neural operator's internal representations, we obtain calibrated uncertainty maps that match deep ensemble across five benchmarks but at a fraction of the training cost of a neural operator. The resulting uncertainty is also spatially interpretable: it concentrates around shock fronts and other regions where the operator's approximation is known to break down, and it remains stable as test geometries drift away from the training distribution. We view this as evidence that the geometry a neural operator learns while solving a PDE already encodes much of what is needed to quantify its own uncertainty, opening a complementary axis to parameter-centric approaches and a natural starting point for UQ on increasingly large geometry-aware operators.

\paragraph{Limitations and future work.} REEF-GP relies on neural operators whose hidden states preserve pointwise spatial correspondence, as in Transolver; extending it to backbones with latent or spectral internal representations (e.g., FNO) is left for future work. Our current experiments involve stationary problems with single-output responses. 
Time-dependent or vector-valued solutions require explicit inclusion of time as an input and multi-output GP extensions. These are natural next steps given the modular post-hoc design of REEF-GP. Finally, although training overhead is an order of magnitude lower than deep ensembles, peak evaluation memory scales as $\mathcal{O}(N_s^2)$ per expert, which is the dominant cost on large 3D meshes. 




\begingroup
\renewcommand{\section}[2]{\subsubsection*{#2}}
\bibliographystyle{unsrt}
\bibliography{references}

@article{RN1881,
   author = {Batlle, Pau and Darcy, Matthieu and Hosseini, Bamdad and Owhadi, Houman},
   title = {Kernel methods are competitive for operator learning},
   journal = {Journal of Computational Physics},
   volume = {496},
   ISSN = {0021-9991},
   DOI = {ARTN 112549
10.1016/j.jcp.2023.112549},
   year = {2024},
   type = {Journal Article}
}

@article{RN270,
   author = {Higdon, Dave and Kennedy, Marc and Cavendish, James C. and Cafeo, John A. and Ryne, Robert D.},
   title = {Combining Field Data and Computer Simulations for Calibration and Prediction},
   journal = {SIAM Journal on Scientific Computing},
   volume = {26},
   number = {2},
   pages = {448-466},
   ISSN = {1064-8275
1095-7197},
   DOI = {10.1137/s1064827503426693},
   url = {<Go to ISI>://WOS:000226823100006},
   year = {2004},
   type = {Journal Article}
}

@article{RN319,
   author = {MacDonald, B. and Ranjan, P. and Chipman, H.},
   title = {GPfit: An R Package for Fitting a Gaussian Process Model to Deterministic Simulator Outputs},
   journal = {Journal of Statistical Software},
   volume = {64},
   number = {12},
   pages = {1-23},
   ISSN = {1548-7660},
   url = {<Go to ISI>://WOS:000352916200001},
   year = {2015},
   type = {Journal Article}
}

@article{RN334,
   author = {Loeppky, J and Bingham, Derek and Welch, W},
   title = {Computer model calibration or tuning in practice},
   journal = {Technometrics, submitted for publication},
   year = {2006},
   type = {Journal Article}
}

@article{lowery2024kernel,
  title={Kernel neural operators (KNOs) for scalable, memory-efficient, geometrically-flexible operator learning},
  author={Lowery, Matthew and Turnage, John and Morrow, Zachary and Jakeman, John D and Narayan, Akil and Zhe, Shandian and Shankar, Varun},
  journal={arXiv preprint arXiv:2407.00809},
  year={2024}
}

@article{batlle2025error,
  title={Error analysis of kernel/GP methods for nonlinear and parametric PDEs},
  author={Batlle, Pau and Chen, Yifan and Hosseini, Bamdad and Owhadi, Houman and Stuart, Andrew M},
  journal={Journal of Computational Physics},
  volume={520},
  pages={113488},
  year={2025},
  publisher={Elsevier}
}

@article{MORA2025117581,
title = {Operator learning with Gaussian processes},
journal = {Computer Methods in Applied Mechanics and Engineering},
volume = {434},
pages = {117581},
year = {2025},
issn = {0045-7825},
doi = {https://doi.org/10.1016/j.cma.2024.117581},
url = {https://www.sciencedirect.com/science/article/pii/S0045782524008351},
author = {Carlos Mora and Amin Yousefpour and Shirin Hosseinmardi and Houman Owhadi and Ramin Bostanabad}
}

@misc{li2021fourierneuraloperatorparametric,
      title={Fourier Neural Operator for Parametric Partial Differential Equations}, 
      author={Zongyi Li and Nikola Kovachki and Kamyar Azizzadenesheli and Burigede Liu and Kaushik Bhattacharya and Andrew Stuart and Anima Anandkumar},
      year={2021},
      eprint={2010.08895},
      archivePrefix={arXiv},
      primaryClass={cs.LG},
      url={https://arxiv.org/abs/2010.08895}, 
}

@misc{matveev2025lightweightdiffusionmultiplieruncertainty,
      title={Light-Weight Diffusion Multiplier and Uncertainty Quantification for Fourier Neural Operators}, 
      author={Albert Matveev and Sanmitra Ghosh and Aamal Hussain and James-Michael Leahy and Michalis Michaelides},
      year={2025},
      eprint={2508.00643},
      archivePrefix={arXiv},
      primaryClass={cs.LG},
      url={https://arxiv.org/abs/2508.00643}, 
}

@article{10.5555/3648699.3649087,
author = {Li, Zongyi and Huang, Daniel Zhengyu and Liu, Burigede and Anandkumar, Anima},
title = {Fourier neural operator with learned deformations for PDEs on general geometries},
year = {2023},
issue_date = {January 2023},
publisher = {JMLR.org},
volume = {24},
number = {1},
issn = {1532-4435},
journal = {J. Mach. Learn. Res.},
month = jan,
articleno = {388},
numpages = {26}
}

@misc{wu2024transolverfasttransformersolver,
      title={Transolver: A Fast Transformer Solver for PDEs on General Geometries}, 
      author={Haixu Wu and Huakun Luo and Haowen Wang and Jianmin Wang and Mingsheng Long},
      year={2024},
      eprint={2402.02366},
      archivePrefix={arXiv},
      primaryClass={cs.LG},
      url={https://arxiv.org/abs/2402.02366}, 
}

@misc{luo2025transolveraccurateneuralsolver,
      title={Transolver++: An Accurate Neural Solver for PDEs on Million-Scale Geometries}, 
      author={Huakun Luo and Haixu Wu and Hang Zhou and Lanxiang Xing and Yichen Di and Jianmin Wang and Mingsheng Long},
      year={2025},
      eprint={2502.02414},
      archivePrefix={arXiv},
      primaryClass={cs.LG},
      url={https://arxiv.org/abs/2502.02414}, 
}

@inproceedings{
li2023geometryinformed,
title={Geometry-Informed Neural Operator for Large-Scale 3D {PDE}s},
author={Zongyi Li and Nikola Borislavov Kovachki and Chris Choy and Boyi Li and Jean Kossaifi and Shourya Prakash Otta and Mohammad Amin Nabian and Maximilian Stadler and Christian Hundt and Kamyar Azizzadenesheli and Anima Anandkumar},
booktitle={Thirty-seventh Conference on Neural Information Processing Systems},
year={2023},
url={https://openreview.net/forum?id=86dXbqT5Ua}
}

@InProceedings{pmlr-v51-wilson16,
  title = 	 {Deep Kernel Learning},
  author = 	 {Wilson, Andrew Gordon and Hu, Zhiting and Salakhutdinov, Ruslan and Xing, Eric P.},
  booktitle = 	 {Proceedings of the 19th International Conference on Artificial Intelligence and Statistics},
  pages = 	 {370--378},
  year = 	 {2016},
  editor = 	 {Gretton, Arthur and Robert, Christian C.},
  volume = 	 {51},
  series = 	 {Proceedings of Machine Learning Research},
  address = 	 {Cadiz, Spain},
  month = 	 {09--11 May},
  publisher =    {PMLR},
  url = 	 {https://proceedings.mlr.press/v51/wilson16.html}
}

@inproceedings{10.5555/3295222.3295349,
author = {Vaswani, Ashish and Shazeer, Noam and Parmar, Niki and Uszkoreit, Jakob and Jones, Llion and Gomez, Aidan N. and Kaiser, \L{}ukasz and Polosukhin, Illia},
title = {Attention is all you need},
year = {2017},
isbn = {9781510860964},
publisher = {Curran Associates Inc.},
address = {Red Hook, NY, USA},
booktitle = {Proceedings of the 31st International Conference on Neural Information Processing Systems},
pages = {6000–6010},
numpages = {11},
location = {Long Beach, California, USA},
series = {NIPS'17}
}

@inproceedings{10.5555/3045390.3045502,
author = {Gal, Yarin and Ghahramani, Zoubin},
title = {Dropout as a Bayesian approximation: representing model uncertainty in deep learning},
year = {2016},
publisher = {JMLR.org},
booktitle = {Proceedings of the 33rd International Conference on International Conference on Machine Learning - Volume 48},
pages = {1050–1059},
numpages = {10},
location = {New York, NY, USA},
series = {ICML'16}
}

@article{10.1162/neco.1992.4.3.415,
    author = {MacKay, David J. C.},
    title = {Bayesian Interpolation},
    journal = {Neural Computation},
    volume = {4},
    number = {3},
    pages = {415-447},
    year = {1992},
    month = {05},
    issn = {0899-7667},
    doi = {10.1162/neco.1992.4.3.415},
    url = {https://doi.org/10.1162/neco.1992.4.3.415},
    eprint = {https://direct.mit.edu/neco/article-pdf/4/3/415/812340/neco.1992.4.3.415.pdf},
}

@inproceedings{
ritter2018a,
title={A Scalable Laplace Approximation for Neural Networks},
author={Hippolyt Ritter and Aleksandar Botev and David Barber},
booktitle={International Conference on Learning Representations},
year={2018},
url={https://openreview.net/forum?id=Skdvd2xAZ},
}

@inproceedings{NEURIPS2021_a7c95857,
 author = {Daxberger, Erik and Kristiadi, Agustinus and Immer, Alexander and Eschenhagen, Runa and Bauer, Matthias and Hennig, Philipp},
 booktitle = {Advances in Neural Information Processing Systems},
 editor = {M. Ranzato and A. Beygelzimer and Y. Dauphin and P.S. Liang and J. Wortman Vaughan},
 pages = {20089--20103},
 publisher = {Curran Associates, Inc.},
 title = {Laplace Redux - Effortless Bayesian Deep Learning},
 url = {https://proceedings.neurips.cc/paper_files/paper/2021/file/a7c9585703d275249f30a088cebba0ad-Paper.pdf},
 volume = {34},
 year = {2021}
}

@article{DBLP:journals/corr/abs-2208-01565,
  author       = {Emilia Magnani and
                  Nicholas Kr{\"{a}}mer and
                  Runa Eschenhagen and
                  Lorenzo Rosasco and
                  Philipp Hennig},
  title        = {Approximate Bayesian Neural Operators: Uncertainty Quantification
                  for Parametric PDEs},
  journal      = {CoRR},
  volume       = {abs/2208.01565},
  year         = {2022},
  url          = {https://doi.org/10.48550/arXiv.2208.01565},
  doi          = {10.48550/ARXIV.2208.01565},
  eprinttype    = {arXiv},
  eprint       = {2208.01565},
  bibsource    = {dblp computer science bibliography, https://dblp.org}
}

@inproceedings{10.5555/3295222.3295387,
author = {Lakshminarayanan, Balaji and Pritzel, Alexander and Blundell, Charles},
title = {Simple and scalable predictive uncertainty estimation using deep ensembles},
year = {2017},
isbn = {9781510860964},
publisher = {Curran Associates Inc.},
address = {Red Hook, NY, USA},
booktitle = {Proceedings of the 31st International Conference on Neural Information Processing Systems},
pages = {6405–6416},
numpages = {12},
location = {Long Beach, California, USA},
series = {NIPS'17}
}

@book{10.7551/mitpress/3206.001.0001,
    author = {Rasmussen, Carl Edward and Williams, Christopher K. I.},
    title = {Gaussian Processes for Machine Learning},
    publisher = {The MIT Press},
    year = {2005},
    month = {11},
    isbn = {9780262256834},
    doi = {10.7551/mitpress/3206.001.0001},
    url = {https://doi.org/10.7551/mitpress/3206.001.0001},
    eprint = {https://direct.mit.edu/book-pdf/2514321/book_9780262256834.pdf},
}

@InProceedings{pmlr-v5-titsias09a,
  title = 	 {Variational Learning of Inducing Variables in Sparse Gaussian Processes},
  author = 	 {Titsias, Michalis},
  booktitle = 	 {Proceedings of the Twelfth International Conference on Artificial Intelligence and Statistics},
  pages = 	 {567--574},
  year = 	 {2009},
  editor = 	 {van Dyk, David and Welling, Max},
  volume = 	 {5},
  series = 	 {Proceedings of Machine Learning Research},
  address = 	 {Hilton Clearwater Beach Resort, Clearwater Beach, Florida USA},
  month = 	 {16--18 Apr},
  publisher =    {PMLR},
  url = 	 {https://proceedings.mlr.press/v5/titsias09a.html}
}

@inproceedings{NIPS2000_19de10ad,
 author = {Williams, Christopher and Seeger, Matthias},
 booktitle = {Advances in Neural Information Processing Systems},
 editor = {T. Leen and T. Dietterich and V. Tresp},
 pages = {},
 publisher = {MIT Press},
 title = {Using the Nystr\"{o}m Method to Speed Up Kernel Machines},
 url = {https://proceedings.neurips.cc/paper_files/paper/2000/file/19de10adbaa1b2ee13f77f679fa1483a-Paper.pdf},
 volume = {13},
 year = {2000}
}

@InProceedings{pmlr-v38-hensman15,
  title = 	 {{Scalable Variational Gaussian Process Classification}},
  author = 	 {Hensman, James and Matthews, Alexander and Ghahramani, Zoubin},
  booktitle = 	 {Proceedings of the Eighteenth International Conference on Artificial Intelligence and Statistics},
  pages = 	 {351--360},
  year = 	 {2015},
  editor = 	 {Lebanon, Guy and Vishwanathan, S. V. N.},
  volume = 	 {38},
  series = 	 {Proceedings of Machine Learning Research},
  address = 	 {San Diego, California, USA},
  month = 	 {09--12 May},
  publisher =    {PMLR},
  url = 	 {https://proceedings.mlr.press/v38/hensman15.html}
}

@inproceedings{10.5555/3045118.3045307,
author = {Wilson, Andrew Gordon and Nickisch, Hannes},
title = {Kernel interpolation for scalable structured Gaussian processes (KISS-GP)},
year = {2015},
publisher = {JMLR.org},
booktitle = {Proceedings of the 32nd International Conference on Machine Learning - Volume 37},
pages = {1775–1784},
numpages = {10},
location = {Lille, France},
series = {ICML'15}
}

@InProceedings{pmlr-v89-zhe19a,
  title = 	 {Scalable High-Order Gaussian Process Regression},
  author =       {Zhe, Shandian and Xing, Wei and Kirby, Robert M.},
  booktitle = 	 {Proceedings of the Twenty-Second International Conference on Artificial Intelligence and Statistics},
  pages = 	 {2611--2620},
  year = 	 {2019},
  editor = 	 {Chaudhuri, Kamalika and Sugiyama, Masashi},
  volume = 	 {89},
  series = 	 {Proceedings of Machine Learning Research},
  month = 	 {16--18 Apr},
  publisher =    {PMLR},
  url = 	 {https://proceedings.mlr.press/v89/zhe19a.html}
}

@article{OWHADI201922,
title = {Kernel Flows: From learning kernels from data into the abyss},
journal = {Journal of Computational Physics},
volume = {389},
pages = {22-47},
year = {2019},
issn = {0021-9991},
doi = {https://doi.org/10.1016/j.jcp.2019.03.040},
url = {https://www.sciencedirect.com/science/article/pii/S0021999119302232},
author = {Houman Owhadi and Gene Ryan Yoo}
}

@article{doi:10.1137/130938633,
author = {Owhadi, Houman and Scovel, Clint and Sullivan, Tim},
title = {On the Brittleness of Bayesian Inference},
journal = {SIAM Review},
volume = {57},
number = {4},
pages = {566-582},
year = {2015},
doi = {10.1137/130938633},
URL = {https://doi.org/10.1137/130938633},
eprint = {https://doi.org/10.1137/130938633}
}

@article{10.1145/3197517.3201325,
author = {Umetani, Nobuyuki and Bickel, Bernd},
title = {Learning three-dimensional flow for interactive aerodynamic design},
year = {2018},
issue_date = {August 2018},
publisher = {Association for Computing Machinery},
address = {New York, NY, USA},
volume = {37},
number = {4},
issn = {0730-0301},
url = {https://doi.org/10.1145/3197517.3201325},
doi = {10.1145/3197517.3201325},
journal = {ACM Trans. Graph.},
month = jul,
articleno = {89},
numpages = {10}
}

@article{JMLR:v24:22-0479,
  author  = {Jeremiah Zhe Liu and Shreyas Padhy and Jie Ren and Zi Lin and Yeming Wen and Ghassen Jerfel and Zachary Nado and Jasper Snoek and Dustin Tran and Balaji Lakshminarayanan},
  title   = {A Simple Approach to Improve Single-Model Deep Uncertainty via Distance-Awareness},
  journal = {Journal of Machine Learning Research},
  year    = {2023},
  volume  = {24},
  number  = {42},
  pages   = {1--63},
  url     = {http://jmlr.org/papers/v24/22-0479.html}
}

@InProceedings{pmlr-v258-jimenez25a,
  title = 	 {Vecchia Gaussian Process Ensembles on Internal Representations of Deep Neural Networks},
  author =       {Jimenez, Felix and Katzfuss, Matthias},
  booktitle = 	 {Proceedings of The 28th International Conference on Artificial Intelligence and Statistics},
  pages = 	 {3403--3411},
  year = 	 {2025},
  editor = 	 {Li, Yingzhen and Mandt, Stephan and Agrawal, Shipra and Khan, Emtiyaz},
  volume = 	 {258},
  series = 	 {Proceedings of Machine Learning Research},
  month = 	 {03--05 May},
  publisher =    {PMLR},
  url = 	 {https://proceedings.mlr.press/v258/jimenez25a.html}
}

@article{JMLR:v24:22-0676,
  author  = {Michail Spitieris and Ingelin Steinsland},
  title   = {Bayesian Calibration of Imperfect Computer Models using Physics-Informed Priors},
  journal = {Journal of Machine Learning Research},
  year    = {2023},
  volume  = {24},
  number  = {108},
  pages   = {1--39},
  url     = {http://jmlr.org/papers/v24/22-0676.html}
}

@article{10.1111/1467-9868.00294,
    author = {Kennedy, Marc C. and O'Hagan, Anthony},
    title = {Bayesian Calibration of Computer Models},
    journal = {Journal of the Royal Statistical Society Series B: Statistical Methodology},
    volume = {63},
    number = {3},
    pages = {425-464},
    year = {2002},
    month = {01},
    issn = {1369-7412},
    doi = {10.1111/1467-9868.00294},
    url = {https://doi.org/10.1111/1467-9868.00294},
    eprint = {https://academic.oup.com/jrsssb/article-pdf/63/3/425/49590547/jrsssb_63_3_425.pdf},
}

@InProceedings{pmlr-v97-rahaman19a,
  title = 	 {On the Spectral Bias of Neural Networks},
  author =       {Rahaman, Nasim and Baratin, Aristide and Arpit, Devansh and Draxler, Felix and Lin, Min and Hamprecht, Fred and Bengio, Yoshua and Courville, Aaron},
  booktitle = 	 {Proceedings of the 36th International Conference on Machine Learning},
  pages = 	 {5301--5310},
  year = 	 {2019},
  editor = 	 {Chaudhuri, Kamalika and Salakhutdinov, Ruslan},
  volume = 	 {97},
  series = 	 {Proceedings of Machine Learning Research},
  month = 	 {09--15 Jun},
  publisher =    {PMLR},
  url = 	 {https://proceedings.mlr.press/v97/rahaman19a.html}
}

@inproceedings{
weber2024uncertainty,
title={Uncertainty Quantification for Fourier Neural Operators},
author={Tobias Weber and Emilia Magnani and Marvin Pf{\"o}rtner and Philipp Hennig},
booktitle={ICLR 2024 Workshop on AI4DifferentialEquations In Science},
year={2024},
url={https://openreview.net/forum?id=knSgoNJcnV}
}

@inproceedings{
magnani2025linearization,
title={Linearization Turns Neural Operators into Function-Valued Gaussian Processes},
author={Emilia Magnani and Marvin Pf{\"o}rtner and Tobias Weber and Philipp Hennig},
booktitle={Forty-second International Conference on Machine Learning},
year={2025},
url={https://openreview.net/forum?id=4Z04wVQ9FY}
}

@inproceedings{lte2024probabilistic,
title={Probabilistic predictions with Fourier neural operators},
author={Christopher B{\"u}lte and Philipp Scholl and Gitta Kutyniok},
booktitle={NeurIPS 2024 Workshop on Bayesian Decision-making and Uncertainty},
year={2024},
url={https://openreview.net/forum?id=orKA6gJwlB}
}

@article{
blte2025probabilistic,
title={Probabilistic neural operators for functional uncertainty quantification},
author={Christopher B{\"u}lte and Philipp Scholl and Gitta Kutyniok},
journal={Transactions on Machine Learning Research},
issn={2835-8856},
year={2025},
url={https://openreview.net/forum?id=gangoPXSRw},
note={}
}

@article{JMLR:v9:vandermaaten08a,
  author  = {Laurens van der Maaten and Geoffrey Hinton},
  title   = {Visualizing Data using t-SNE},
  journal = {Journal of Machine Learning Research},
  year    = {2008},
  volume  = {9},
  number  = {86},
  pages   = {2579--2605},
  url     = {http://jmlr.org/papers/v9/vandermaaten08a.html}
}

@misc{zhou2026transolver3scalingtransformersolvers,
      title={Transolver-3: Scaling Up Transformer Solvers to Industrial-Scale Geometries}, 
      author={Hang Zhou and Haixu Wu and Haonan Shangguan and Yuezhou Ma and Huikun Weng and Jianmin Wang and Mingsheng Long},
      year={2026},
      eprint={2602.04940},
      archivePrefix={arXiv},
      primaryClass={cs.LG},
      url={https://arxiv.org/abs/2602.04940}, 
}

@inbook{10.5555/3454287.3455466,
author = {Maddox, Wesley J. and Garipov, Timur and Izmailov, Pavel and Vetrov, Dmitry and Wilson, Andrew Gordon},
title = {A simple baseline for Bayesian uncertainty in deep learning},
year = {2019},
publisher = {Curran Associates Inc.},
address = {Red Hook, NY, USA},
booktitle = {Proceedings of the 33rd International Conference on Neural Information Processing Systems},
articleno = {1179},
numpages = {12}
}

@misc{pathak2022fourcastnetglobaldatadrivenhighresolution,
      title={FourCastNet: A Global Data-driven High-resolution Weather Model using Adaptive Fourier Neural Operators}, 
      author={Jaideep Pathak and Shashank Subramanian and Peter Harrington and Sanjeev Raja and Ashesh Chattopadhyay and Morteza Mardani and Thorsten Kurth and David Hall and Zongyi Li and Kamyar Azizzadenesheli and Pedram Hassanzadeh and Karthik Kashinath and Animashree Anandkumar},
      year={2022},
      eprint={2202.11214},
      archivePrefix={arXiv},
      primaryClass={physics.ao-ph},
      url={https://arxiv.org/abs/2202.11214}, 
}

@inproceedings{
ayhan2018testtime,
title={Test-time Data Augmentation for Estimation of Heteroscedastic Aleatoric Uncertainty in Deep Neural Networks},
author={Murat Seckin Ayhan and Philipp Berens},
booktitle={Medical Imaging with Deep Learning},
year={2018},
url={https://openreview.net/forum?id=rJZz-knjz}
}

@inproceedings{
skean2025layer,
title={Layer by Layer: Uncovering Hidden Representations in Language Models},
author={Oscar Skean and Md Rifat Arefin and Dan Zhao and Niket Nikul Patel and Jalal Naghiyev and Yann LeCun and Ravid Shwartz-Ziv},
booktitle={Forty-second International Conference on Machine Learning},
year={2025},
url={https://openreview.net/forum?id=WGXb7UdvTX}
}

@misc{he2015deepresiduallearningimage,
      title={Deep Residual Learning for Image Recognition}, 
      author={Kaiming He and Xiangyu Zhang and Shaoqing Ren and Jian Sun},
      year={2015},
      eprint={1512.03385},
      archivePrefix={arXiv},
      primaryClass={cs.CV},
      url={https://arxiv.org/abs/1512.03385}, 
}

@misc{chang2015shapenetinformationrich3dmodel,
      title={ShapeNet: An Information-Rich 3D Model Repository}, 
      author={Angel X. Chang and Thomas Funkhouser and Leonidas Guibas and Pat Hanrahan and Qixing Huang and Zimo Li and Silvio Savarese and Manolis Savva and Shuran Song and Hao Su and Jianxiong Xiao and Li Yi and Fisher Yu},
      year={2015},
      eprint={1512.03012},
      archivePrefix={arXiv},
      primaryClass={cs.GR},
      url={https://arxiv.org/abs/1512.03012}, 
}

@misc{loshchilov2019decoupledweightdecayregularization,
      title={Decoupled Weight Decay Regularization}, 
      author={Ilya Loshchilov and Frank Hutter},
      year={2019},
      eprint={1711.05101},
      archivePrefix={arXiv},
      primaryClass={cs.LG},
      url={https://arxiv.org/abs/1711.05101}, 
}

@misc{hendrycks2023gaussianerrorlinearunits,
      title={Gaussian Error Linear Units (GELUs)}, 
      author={Dan Hendrycks and Kevin Gimpel},
      year={2023},
      eprint={1606.08415},
      archivePrefix={arXiv},
      primaryClass={cs.LG},
      url={https://arxiv.org/abs/1606.08415}, 
}

@article{YOUSEFPOUR2024103686,
title = {GP+: A Python library for kernel-based learning via Gaussian processes},
journal = {Advances in Engineering Software},
volume = {195},
pages = {103686},
year = {2024},
issn = {0965-9978},
doi = {https://doi.org/10.1016/j.advengsoft.2024.103686},
url = {https://www.sciencedirect.com/science/article/pii/S0965997824000930},
author = {Amin Yousefpour and Zahra Zanjani Foumani and Mehdi Shishehbor and Carlos Mora and Ramin Bostanabad}
}

@inproceedings{
musekamp2025active,
title={Active Learning for Neural {PDE} Solvers},
author={Daniel Musekamp and Marimuthu Kalimuthu and David Holzm{\"u}ller and Makoto Takamoto and Mathias Niepert},
booktitle={The Thirteenth International Conference on Learning Representations},
year={2025},
url={https://openreview.net/forum?id=x4ZmQaumRg}
}

@misc{cao2015generalizedproductexpertsautomatic,
      title={Generalized Product of Experts for Automatic and Principled Fusion of Gaussian Process Predictions}, 
      author={Yanshuai Cao and David J. Fleet},
      year={2015},
      eprint={1410.7827},
      archivePrefix={arXiv},
      primaryClass={cs.LG},
      url={https://arxiv.org/abs/1410.7827}, 
}

@article{DBLP:journals/corr/abs-1905-00414,
  author       = {Simon Kornblith and
                  Mohammad Norouzi and
                  Honglak Lee and
                  Geoffrey E. Hinton},
  title        = {Similarity of Neural Network Representations Revisited},
  journal      = {CoRR},
  volume       = {abs/1905.00414},
  year         = {2019},
  url          = {http://arxiv.org/abs/1905.00414},
  eprinttype   = {arXiv},
  eprint       = {1905.00414},
  bibsource    = {dblp computer science bibliography, https://dblp.org}
}

@article{JMLR:v23:20-1365,
  author  = {Hao Chen and Lili Zheng and Raed Al Kontar and Garvesh Raskutti},
  title   = {Gaussian Process Parameter Estimation Using Mini-batch Stochastic Gradient Descent: Convergence Guarantees and Empirical Benefits},
  journal = {Journal of Machine Learning Research},
  year    = {2022},
  volume  = {23},
  number  = {227},
  pages   = {1--59},
  url     = {http://jmlr.org/papers/v23/20-1365.html}
}

@misc{yang2024minibatchmethodsolvingnonlinear,
      title={A Mini-Batch Method for Solving Nonlinear PDEs with Gaussian Processes}, 
      author={Xianjin Yang and Houman Owhadi},
      year={2024},
      eprint={2306.00307},
      archivePrefix={arXiv},
      primaryClass={math.NA},
      url={https://arxiv.org/abs/2306.00307}, 
}

@article{PSAROS2023111902,
    title = {Uncertainty quantification in scientific machine learning: Methods, metrics, and comparisons},
    journal = {Journal of Computational Physics},
    volume = {477},
    pages = {111902},
    year = {2023},
    issn = {0021-9991},
    doi = {https://doi.org/10.1016/j.jcp.2022.111902},
    url = {https://www.sciencedirect.com/science/article/pii/S0021999122009652},
    author = {Apostolos F. Psaros and Xuhui Meng and Zongren Zou and Ling Guo and George Em Karniadakis},
}

@inproceedings{
rahman2024pretraining,
title={Pretraining Codomain Attention Neural Operators for Solving Multiphysics {PDE}s},
author={Md Ashiqur Rahman and Robert Joseph George and Mogab Elleithy and Daniel Leibovici and Zongyi Li and Boris Bonev and Colin White and Julius Berner and Raymond A. Yeh and Jean Kossaifi and Kamyar Azizzadenesheli and Anima Anandkumar},
booktitle={The Thirty-eighth Annual Conference on Neural Information Processing Systems},
year={2024},
url={https://openreview.net/forum?id=wSpIdUXZYX}
}

@inproceedings{10.5555/3618408.3618525,
author = {Bonev, Boris and Kurth, Thorsten and Hundt, Christian and Pathak, Jaideep and Baust, Maximilian and Kashinath, Karthik and Anandkumar, Anima},
title = {Spherical Fourier neural operators: learning stable dynamics on the sphere},
year = {2023},
publisher = {JMLR.org},
booktitle = {Proceedings of the 40th International Conference on Machine Learning},
articleno = {117},
numpages = {18},
location = {Honolulu, Hawaii, USA},
series = {ICML'23}
}

@misc{khorrami2025physicsencodedfourierneuraloperator,
      title={A physics-encoded Fourier neural operator approach for surrogate modeling of divergence-free stress fields in solids}, 
      author={Mohammad S. Khorrami and Pawan Goyal and Jaber R. Mianroodi and Bob Svendsen and Peter Benner and Dierk Raabe},
      year={2025},
      eprint={2408.15408},
      archivePrefix={arXiv},
      primaryClass={cs.CE},
      url={https://arxiv.org/abs/2408.15408}, 
}

@misc{mouli2024usinguncertaintyquantificationcharacterize,
      title={Using Uncertainty Quantification to Characterize and Improve Out-of-Domain Learning for PDEs}, 
      author={S. Chandra Mouli and Danielle C. Maddix and Shima Alizadeh and Gaurav Gupta and Andrew Stuart and Michael W. Mahoney and Yuyang Wang},
      year={2024},
      eprint={2403.10642},
      archivePrefix={arXiv},
      primaryClass={cs.LG},
      url={https://arxiv.org/abs/2403.10642}, 
}

@misc{jeyaraj2025neuraloperatorbasedhybrid,
      title={A Neural Operator based Hybrid Microscale Model for Multiscale Simulation of Rate-Dependent Materials}, 
      author={Dhananjeyan Jeyaraj and Hamidreza Eivazi and Jendrik-Alexander Tröger and Stefan Wittek and Stefan Hartmann and Andreas Rausch},
      year={2025},
      eprint={2506.16918},
      archivePrefix={arXiv},
      primaryClass={physics.comp-ph},
      url={https://arxiv.org/abs/2506.16918}, 
}

@misc{gopakumar2023fourierneuraloperatorplasma,
      title={Fourier Neural Operator for Plasma Modelling}, 
      author={Vignesh Gopakumar and Stanislas Pamela and Lorenzo Zanisi and Zongyi Li and Anima Anandkumar and MAST Team},
      year={2023},
      eprint={2302.06542},
      archivePrefix={arXiv},
      primaryClass={physics.plasm-ph},
      url={https://arxiv.org/abs/2302.06542}, 
}

@misc{carey2025neuraloperatorsurrogatemodels,
      title={Neural operator surrogate models of plasma edge simulations: feasibility and data efficiency}, 
      author={N. Carey and L. Zanisi and S. Pamela and V. Gopakumar and J. Omotani and J. Buchanan and J. Brandstetter and F. Paischer and G. Galletti and P. Setinek},
      year={2025},
      eprint={2502.17386},
      archivePrefix={arXiv},
      primaryClass={physics.plasm-ph},
      url={https://arxiv.org/abs/2502.17386}, 
}
\endgroup

\clearpage

\appendix

\section{Operator Learning with Gaussian Processes}\label{appendix:gps-for-operator-learning}

We adopt the \textit{functional regression} perspective of \cite{MORA2025117581} as opposed to the operator-valued approach of \cite{RN1881} (we note that neither works study the UQ properties of GPs in the context of operator learning). Specifically, instead of learning the map $\mathcal{G}^\dagger$ directly (which outputs an infinite-dimensional function), we learn the evaluation functional associated with the operator.
Formally, we define a real-valued bilinear form $\tilde{\mathcal{G}}^\dagger$ acting on the input space and the dual space of the output:
\begin{equation}
    \tilde{\mathcal{G}}^\dagger: (\mathcal{U} \times \mathcal{A}) \times \mathcal{V}^* \to \mathbb{R}, \quad ((u, a), \varphi) \mapsto \langle \varphi, \mathcal{G}^\dagger(u, a) \rangle_{\mathcal{V}^* \times \mathcal{V}},
\end{equation}
where $\mathcal{V}^*$ is the dual space of $\mathcal{V}$ and $\langle \cdot, \cdot \rangle$ denotes the duality pairing. In particular, by choosing $\varphi$ to be the Dirac delta functional $\delta_\xb$ centered at $\xb \in D_a$, the pairing recovers the pointwise evaluation of the solution field:
\begin{equation}
    \tilde{\mathcal{G}}^\dagger((u, a), \delta_\xb) = \langle \delta_\xb, v \rangle = v(\xb).
\end{equation}
This transformation effectively converts the operator learning problem into a scalar regression task, where the spatial coordinate $\xb$ becomes an input variable. The process can be extended to vector-valued cases by either repeating the above process or solving a multi-output regression problem. 

In practice, we only have access to finite discretizations. Let $\mathcal{E}_u: \mathcal{U} \to \mathbb{R}^{d_u}$ be a bounded linear observation operator that extracts discrete $d_u$ dimensional features from the continuous input function (e.g., sensor measurements or nodes of a mesh in computer simulations) such that $\mathbf{u} = \mathcal{E}_u(u)$. Similarly, let $\mathcal{E}_a: \mathcal{A} \to \mathbb{R}^{d_a}$ denote a discretization operator that produces the discrete $d_a$-dimensional representation of the geometry such that $\mathbf{a} = \mathcal{E}_a(a)$. Consequently, the infinite-dimensional functional $\tilde{\mathcal{G}}^\dagger$ is approximated by a finite-dimensional surrogate:
\begin{equation}
    f: \mathbb{R}^{d_u} \times \mathbb{R}^{d_a} \times \Omega \to \mathbb{R}, \quad (\inu, \ingeom, \xb) \mapsto v(\xb). \tag{\ref{eq:gp_for_ol}}
\end{equation}

This formulation allows us to treat the triplet $\mathbf{z} = (\inu, \ingeom, \xb)$ as the joint input to a regression model. By placing a GP prior over $f$, i.e., $f(\mathbf{z}) \sim \mathcal{GP}(m(\mathbf{z}), k(\mathbf{z}, \mathbf{z}'))$, we can infer the solution value $v(\xb)$ at any query point $\xb$ within the irregular domain $D_a$, while naturally obtaining uncertainty estimates via the GP posterior.

\section{Implementation Details of REEF-GP}\label{appendix:implementation-details}

This appendix details our design choices as well as training and inference algorithms that can be used to instantiate REEF-GP and reproduce our results. GitHub link will be available after review.

\subsection{Layer Selection}\label{appendix:layer_selection}

\begin{figure}[!t] 
    \centering 
        \includegraphics[width=\linewidth]{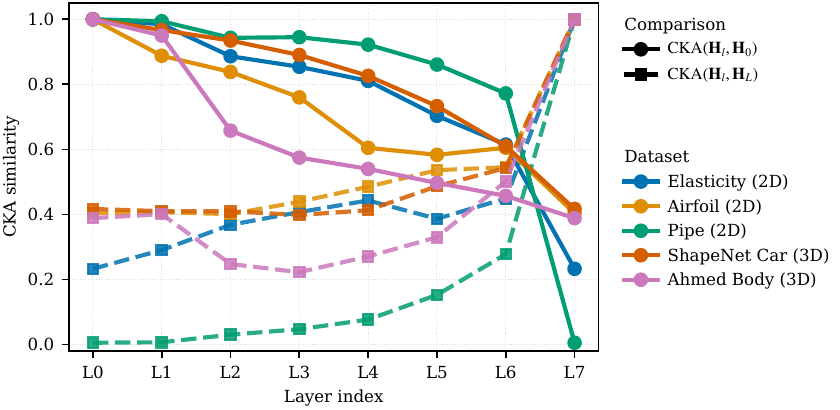} 
    \caption{\textbf{Layer-wise CKA similarity across benchmarks.} 
    For each candidate Transolver layer $\mathbf{H}_l$, we plot CKA similarity to the input layer $\mathbf{H}_0$ (solid) and to the output layer $\mathbf{H}_L$ (dashed) as a function of depth. Intermediate layers are jointly distinct from both endpoints and our middle-layer selection chooses the maximally distinct one.} 
    \label{fig:cka_layer_selection} 
\end{figure}

REEF-GP requires an augmented state $\augz$ that aggregates hidden features $\mathbf{h}_l(\mathbf{u},\mathbf{a},\xb)$ from $L'$ selected layers of a frozen base neural operator (Transolver in this case). Naively concatenating all $L=8$ Transolver layers would produce a very high-dimensional representation that increases the costs while providing negligible gains as these representations have redundancies. Internal Transolver layers exhibit strong feature correlation so consecutive blocks contribute overlapping information, see Figure \ref{fig:cka_layer_selection}. We therefore select a small subset of layers that jointly span the network's representational hierarchy: an early layer (preserving raw geometric structure), a late layer (encoding the predicted physics), and a middle layer that is maximally distinct from both. Below, we provide a heuristic for choosing the middle layer which has been used in all our studies. Unlike DVE, we observe that not all layers are needed, see ablation studies in Table \ref{tab:ablation_layer_selection}.

\paragraph{CKA-based middle layer selection.} 
For the middle layer, we use a centered kernel alignment (CKA) \cite{DBLP:journals/corr/abs-1905-00414} criterion to identify the block whose representation is jointly most distinct from the first and last 
layers. We pass a batch of $B$ samples through the frozen Transolver and extract the hidden state at each candidate layer $l$. Flattening across the batch and spatial dimensions yields $\mathbf{H}_l \in \mathbb{R}^{(B \cdot N) \times D_h}$. We randomly subsample $P_s = 5{,}000$ rows and center each subsampled tensor before computing CKA. For two centered matrices $\mathbf{X}$ and $\mathbf{Y}$, linear CKA is:
\begin{equation}
    \text{CKA}(\mathbf{X}, \mathbf{Y}) = 
    \frac{\|\mathbf{Y}^\top \mathbf{X}\|_F^2}
         {\|\mathbf{X}^\top \mathbf{X}\|_F \cdot \|\mathbf{Y}^\top \mathbf{Y}\|_F}.
\end{equation}
The middle layer $l^*$ is then chosen to minimize the combined similarity to the boundary layers:
\begin{equation}
    l^* = \arg\min_{l \in \{1, \dots, L-1\}} 
    \big[\, \text{CKA}(\mathbf{H}_l, \mathbf{H}_0) 
    + \text{CKA}(\mathbf{H}_l, \mathbf{H}_L) \,\big].
\end{equation}
This procedure runs once per dataset and seed, takes negligible time 
relative to GP training, and selects three layers in total: 
$\{L_0, l^*, L_{\text{last}}\}$. Their outputs are concatenated along the feature 
dimension to form $\mathbf{h}(\mathbf{u}, \mathbf{a}, \xb)$. Hidden 
dimensions vary per dataset (Table~\ref{tab:transolver_hyperparams}), so 
the resulting augmented state has dimension $3 D_h$.

\subsection{Deep Kernel Design}\label{appendix:kernel_construction}
The GP prior on $f$ in Equation \ref{eq:gp_for_ol} has a deep kernel that operates on three input types (spatial coordinates $\xb$, discretized input function $\inu$, and the selected embeddings $\mathbf{h}$ from Transolver). 

\paragraph{Latent feature projection.} We project the high-dimensional augmented state $\mathbf{h}$ through a learned MLP $\rho: \mathbb{R}^{3D_h} \to \mathbb{R}^{D_l}$ that 
mixes the three layers and projects them into a compact latent space for the kernel. The network consists of two spectrally normalized linear layers with a ReLU activation:
\begin{equation}
    \rho(\mathbf{h}) = W_2 \cdot \text{ReLU}(W_1 \mathbf{h})
\end{equation}
Spectral normalization bounds the Lipschitz constant of $\rho$, preventing the projection from collapsing the input geometry manifold and preserving the latent distance metric for the kernel~\cite{JMLR:v24:22-0479}.

\paragraph{Kernel composition.} The composite kernel is a product of three base kernels:
\begin{equation}
    k_{\phib}(\augz, \augz') = \sigma^2 \cdot k_{\text{space}}(\xb, \xb') \cdot k_{\text{fun}}(\inu, \inu') \cdot k_{\text{latent}}(\rho(\mathbf{h}), \rho(\mathbf{h}')),
\end{equation}
where $\sigma^2$ is a learned global outputscale, $k_{\text{space}}$ acts on physical coordinates ($d = 2$ or $3$), $k_{\text{fun}}$ acts on the discretized input function when present (e.g., tension traction on the top edge for \emph{Elasticity}), and $k_{\text{latent}}$ acts on the projected augmented state. The function-dimension factor $k_{\text{fun}}$ is omitted when no input function is present. We use radial basis function (RBF) with automatic relevance determination (ARD) lengthscales as the base kernel across all the datasets. Additionally, as recommended in \cite{RN319,YOUSEFPOUR2024103686}, change of variable is used so that the lengthscales $w$ are all optimized in the $\log_{10}$-space, e.g., 
$k_{\text{space}}(\xb, \xb') = \text{exp}\left( -\sum_{r=1}^{d} 10^{w_r}(x_r - x'_r)^2 \right)$.
To ensure numerical stability, we restrict the search interval to $[-5, 3]$.

\paragraph{Hyperprior Design.} 
To avoid overfitting, we place Gaussian priors on the lengthscales and log-process variance:
\begin{align*}
    w_{\text{space}} &\sim \mathcal{N}(-1.5,\, 0.5^2), \\
    w_{\text{fun}} &\sim \mathcal{N}(0.0,\, 0.5^2), \\
    w_{\text{latent}} &\sim \mathcal{N}(-1.0,\, 0.5^2), \\
    \log \sigma^2 &\sim \mathcal{N}(0.0, 0.5^2).
\end{align*}
These priors are fixed across all experiments. No prior is used for the parameters of $\rho(\cdot)$.

\subsection{Heteroscedastic Noise Likelihood}\label{appendix:noise_model}

The structural error $\varepsilon$ and the observational noise $\epsilon$ are functionally indistinguishable from the GP’s perspective as both contribute to the diagonal of the covariance matrix. They are jointly absorbed into a single input-dependent noise variance $\sigma_{\psib}^2(\augz) + \sigma_n^2$ parametrized by an MLP as:
\begin{equation}
    \sigma_{\psib}^2(\augz) + \sigma_n^2 = \text{Softplus}\big( g_{\psib}(\augz) \big) + \sigma_{\min}^2,
\end{equation}
where $g_{\psib}$ is a 3-layer MLP with hidden dimensions $[64, 32]$, output dimension of 1, GELU activations, layer normalization at the input, and spectral normalization on each linear layer. We set $\sigma_{\min}^2 = 10^{-3}$ to ensure numerical conditioning of the covariance matrix. The final linear layer's bias is initialized to $-3.0$ so that $\sigma_{\psib}^2$ starts at a small value. The MLP receives the same input representation used by the kernel $\{\xb,\inu,\mathbf{h}\}$ and it returns a diagonal noise covariance:
\begin{equation}
    p(v(\xb) \mid f(\augz)) = \mathcal{N}\big(f(\augz), \; \sigma_{\psib}^2(\augz) + \sigma_n^2 \big).
\end{equation}

\subsection{Training}\label{appendix:training_algorithm}

\begin{figure}[h] 
    \centering 
        \includegraphics[width=\linewidth]{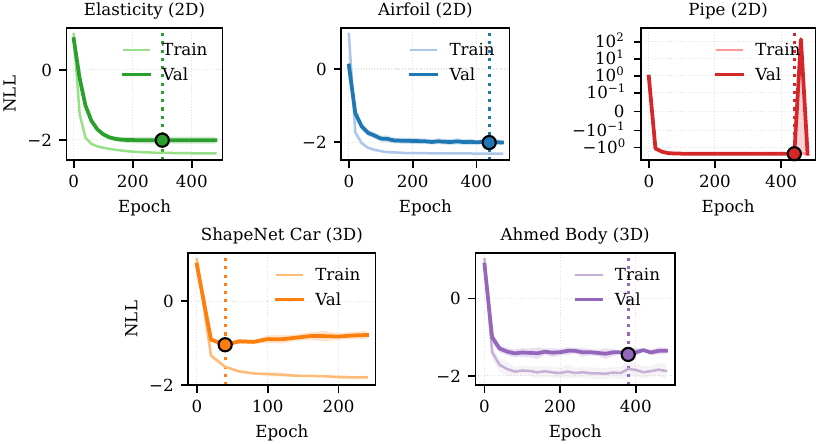} 
    \caption{\textbf{NLL loss curves across benchmarks for five seeds.} 
     Solid lines represent the mean NLL, with shaded regions indicating one standard deviation. To prevent overfitting of the GP hyperparameters, we continuously track the validation loss and employ early stopping. The optimal stopping point, where validation loss reaches its minimum, is marked by the dotted vertical line and point in each subplot.} 
    \label{fig:loss_curves} 
\end{figure}

The full functional regression problem has $MN \in [5 \times 10^5, 5 \times 10^6]$ training points across our benchmarks, making exact GP training intractable. To circumvent this, we draw inspiration from Kernel Flows (KF) \cite{OWHADI201922} and employ stochastic subsampling. While KF loss typically minimizes an RKHS stability metric, we directly minimize the mini-batch negative marginal log-likelihood (NLL) which, unlike KF loss, also prioritizes uncertainty calibration. Although optimizing on mini-batches yields a biased estimator of the full-data NLL, recent works have successfully leveraged random mini-batches \cite{OWHADI201922, JMLR:v23:20-1365, yang2024minibatchmethodsolvingnonlinear} and our empirical studies demonstrate similar trends, see Figure \ref{fig:loss_curves}. We train using a fixed max-iteration budget and the configurations detailed in Table \ref{tab:our_method_hyperparams}.

\begin{table}[h]
\centering
\caption{\textbf{Hyperparameter configurations for our method across benchmarks.} 
Almost the same configuration is used for all five datasets with minimal per-task tuning, demonstrating robustness across varied geometries and physical regimes.}
\label{tab:our_method_hyperparams}
\begin{tabular}{lccccc}
\toprule
\textbf{Hyperparameter} & \textbf{Elasticity} & \textbf{Airfoil} & \textbf{Pipe} & \textbf{ShapeNet Car} & \textbf{Ahmed Body} \\
\midrule
Latent dim $D_l$ & 64 & 64 & 64 & 64 & 64 \\
Spatial kernel $k_{\text{space}}$ & RBF & RBF & RBF & RBF & RBF \\
Input function kernel $k_{\text{fun}}$ & RBF & -- & -- & -- & RBF \\
Latent kernel $k_{\text{latent}}$ & RBF & RBF & RBF & RBF & RBF \\
Subset size $N_s$ & 5{,}000 & 5{,}000 & 5{,}000 & 5{,}000 & 5{,}000 \\
Number of experts $K$ & 5 & 5 & 5 & 5 & 5 \\
Eval chunk size $N_q$ & 10{,}000 & 10{,}000 & 10{,}000 & 10{,}000 & 10{,}000 \\
\midrule
Max iterations & 500 & 500 & 500 & 500 & 500 \\
NN learning rate $\eta_{\text{nn}}$ & $2\times10^{-3}$ & $2\times10^{-3}$ & $2\times10^{-4}$ & $2\times10^{-3}$ & $2\times10^{-3}$ \\
GP learning rate $\eta_{\text{gp}}$ & $5\times10^{-2}$ & $5\times10^{-2}$ & $1\times10^{-3}$ & $5\times10^{-2}$ & $5\times10^{-2}$ \\
Weight decay (NN) & $1\times10^{-4}$ & $1\times10^{-4}$ & $1\times10^{-4}$ & $1\times10^{-4}$ & $1\times10^{-4}$ \\
\bottomrule
\end{tabular}
\end{table}

\paragraph{Numerical stability.} 
GP marginal likelihood and predictive computations are performed using 
exact Cholesky decomposition (jitter $10^{-6}$) when the kernel remains positive definite, and we fall back to conjugate gradient (CG) approximation if it becomes mildly ill-conditioned during training. With a support set of $N_s = 5000$ points, Cholesky 
is computationally feasible (\(O(N_s^3)\) cost). We additionally clip gradients to a maximum norm of $10$ before each optimizer step.

\paragraph{Reproducibility details.} 
 Base Transolver models are previously trained using NVIDIA A100 GPUs (40~GB). All post-hoc UQ method fitting and inference for both our method and the baselines are performed on a single workstation equipped with an NVIDIA RTX 6000 Ada Generation GPU (48~GB). This split reflects realistic deployment: the deterministic operator is trained once leveraging HPC resources, while the post-hoc UQ component is fit and queried on a workstation. All experiments use \texttt{float32} precision. Baselines for MC Dropout, PNO, Perturbation, LUNO-LA, and DVE-spatial share the same Transolver backbone described in Appendix~\ref{appendix:base_no}. Random seeds $\{0, 1, 2, 3, 4\}$ are used across all experiments.

\subsection{Inference}\label{appendix:inference_algorithm}

At inference time, conditioning on the full support set of $MN$ training points remains computationally prohibitive. We employ a uniformly weighted generalized Product of Experts (gPoE) \cite{cao2015generalizedproductexpertsautomatic}: $K$ random support subsets of size $N_s$ condition $K$ GP experts, each producing a posterior predictive mean $\bar{\delta}_k(\augz_*)$ and variance $\sigma_k^2(\augz_*)$ at a query point $\augz_* = (\inu_*, \xb_*, \mathbf{h}(\inu_*, \ingeom_*, \xb_*))$. Unlike mixture aggregation, gPoE combines experts multiplicatively in precision space, yielding a closed-form Gaussian:
\begin{equation}
    \frac{1}{\sigma_*^2(\augz_*)} = \frac{1}{K} \sum_{k=1}^K \frac{1}{\sigma_k^2(\augz_*)}, \quad\quad \bar{\delta}_*(\augz_*) = \sigma_*^2(\augz_*) \cdot \frac{1}{K} \sum_{k=1}^K \frac{\bar{\delta}_k(\augz_*)}{\sigma_k^2(\augz_*)}.
\end{equation}

The aggregated discrepancy is added back to the frozen base operator to form the final predictive distribution:
\begin{equation}
    v(\xb_*) \sim \mathcal{N}\Big(\mathcal{G}_\theta(\inu_*,\ingeom_*)(\xb_*) + \bar{\delta}_*(\augz_*), \; \sigma_*^2(\augz_*) \Big).
\end{equation}
This formulation bounds the test-time memory footprint strictly to $\mathcal{O}(N_s^2)$ per expert, circumventing the $\mathcal{O}((MN)^2)$ requirement of the full support set, while remaining exact within each expert's local approximation. The GP experts use the same hyperparameters and only differ in their conditioning data.

\paragraph{Memory complexity at inference.} The dominant memory cost is the $\mathcal{O}(N_s^2)$ kernel matrix stored once per expert (re-used across all query batches). Query points are processed in chunks of $N_q=10{,}000$ to bound peak memory; this gives a worst-case per-batch peak of $\mathcal{O}(N_s \cdot N_q)$.

\section{Deformation Flow of the Geometry}\label{appendix:kernel_flows}

In Section~\ref{sec:kernel} we introduce a deep kernel formulation that relies on the frozen internal representations of a pretrained neural operator rather than learning a feature map from scratch. This design connects two established lines of work: deep kernel learning (DKL) \cite{pmlr-v51-wilson16}, which composes a stationary kernel with a learned feature extractor, and KF \cite{OWHADI201922}, which constructs data-dependent kernels through iterative, flow-based deformations of the input space. Mechanically, REEF-GP is closest to DKL but conceptually several of our design choices are inspired by KF. The remainder of this appendix formalizes these connections and empirically analyzes the layer-wise geometry deformation of the Transolver backbone.

\paragraph{Residual Networks as KF.}
KF constructs a deep kernel by evaluating a base kernel $K_1$ on a deformed input space. For two input states $\mathbf{z}$ and $\mathbf{z}'$, the kernel takes the form $K(\mathbf{z}, \mathbf{z}') = K_1(F_L(\mathbf{z}), F_L(\mathbf{z}'))$ where $F_L$ is a flow map learned via a sequence of incremental deformations: \begin{equation}
F_{l+1}(\mathbf{z}) = F_l(\mathbf{z}) + \tau\, G_{l+1}(F_l(\mathbf{z})),
\end{equation}
with $\tau$ acting as a discrete step size and $G_{l+1}$ a transformation in the span of kernel evaluations. This incremental update mirrors the residual blocks within Transolver \cite{wu2024transolverfasttransformersolver} and motivates us to leverage its internal geometry-aware representations.

Transolver is built upon transformer blocks \cite{10.5555/3295222.3295349} with residual connections \cite{he2015deepresiduallearningimage}. Let $\mathbf{h}_l$ denote the latent state at layer $l$ corresponding to a spatial coordinate $\xb$. The layer-wise update reads:
\begin{equation}
\mathbf{h}_{l+1} = \mathbf{h}_l + \mathcal{F}_{\text{attn}}(\mathbf{h}_l)
+ \mathcal{F}_{\text{MLP}}(\mathbf{h}_l + \mathcal{F}_{\text{attn}}(\mathbf{h}_l)),
\end{equation}
where $\mathcal{F}_{\text{attn}}$ is the Physics-Attention mechanism. Structurally, Transolver executes a discrete flow mapping that deforms the original coordinate and input space $\mathcal{U} \times \mathcal{A} \times \Omega$ into a structured latent manifold. We do not claim that Transolver is a KF model (it is trained via empirical risk minimization against the PDE solution, not the KF stability loss) but the layer-wise update has the same residual-flow geometry, which is what suggests that the intermediate hidden states carry useful information beyond the final embedding alone.

\begin{figure}[h]
    \centering
        \includegraphics[width=\textwidth]{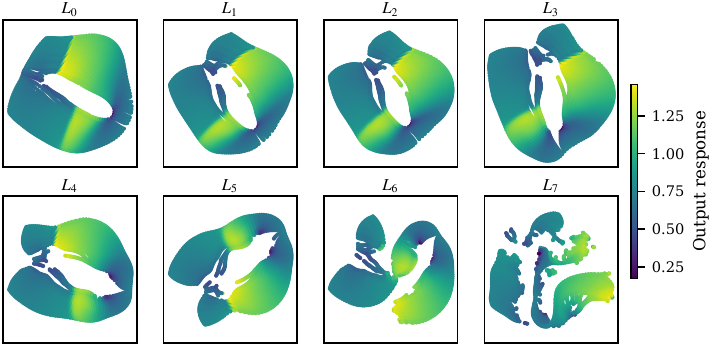}
    \caption{\textbf{Deformation flow of the geometry for an airfoil sample with shock waves above and below the airfoil.} Each panel shows the two-dimensional t-SNE representations of the pointwise embeddings of the initial geometry at each of the 8 layers of the Transolver backbone. Across depth, the network progressively deforms the original geometry into a latent manifold, supporting the view that intermediate layers consistently encode geometry-aware structure.}
    \label{fig:transolver_tsne_0}
\end{figure}

\begin{figure}[!t]
    \centering
        \includegraphics[width=\textwidth]{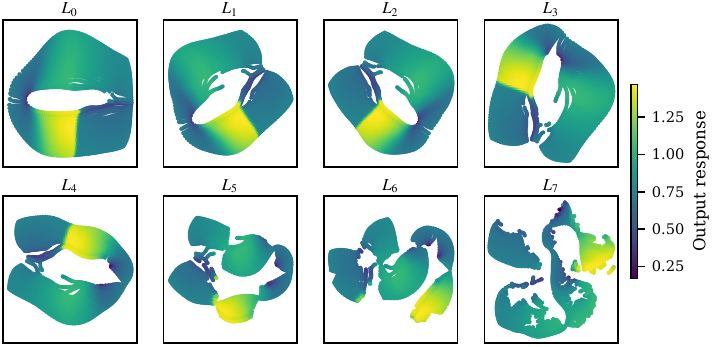}
    \caption{\textbf{Deformation flow of the geometry for an airfoil sample with a shock wave below the airfoil.} Each panel shows the two-dimensional t-SNE representations of the pointwise embeddings of the initial geometry at each of the 8 layers of the Transolver backbone.}
    \label{fig:transolver_tsne_1}
\end{figure}

Figures~\ref{fig:transolver_tsne_0} and \ref{fig:transolver_tsne_1} show the t-SNE representation of the internal embeddings of all the Transolver layers for two airfoil samples. The network learns to deform the geometry into latent representations that resolve the physics of the problem.


\begin{figure}[!b] 
    \centering 
    
    \begin{subfigure}{\textwidth}
        \centering
        \includegraphics[width=\linewidth]{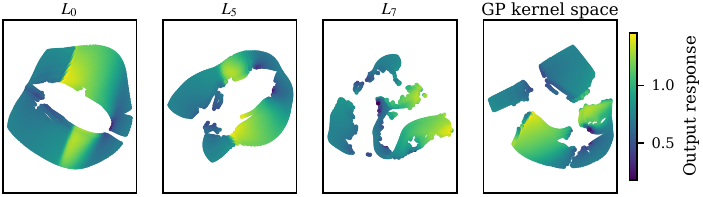}
    \end{subfigure}
    
    \vspace{1em} 
    
    \begin{subfigure}{\textwidth}
        \centering
        \includegraphics[width=\linewidth]{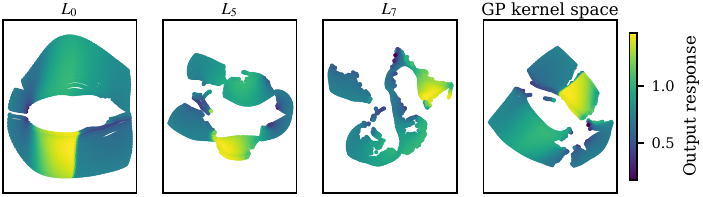} 
    \end{subfigure}
    
    \caption{\textbf{Feature map from Transolver embeddings to the GP kernel space.} Top: airfoil sample with shock waves at both upper and lower surfaces. Bottom: airfoil sample with shock wave at the lower surface. The first three columns show t-SNE projections of the three CKA-selected Transolver layers used to construct the augmented state $\mathbf{h}(\mathbf{u},\mathbf{a},\xb)$; the rightmost column shows the embedding after the spectral-normalized projection $\rho(\cdot)$ used inside the kernel. The GP kernel space tears precisely at the shock wave region, indicating that the projection reshapes the operator's latent geometry into a metric space where uncertainty can localize around physically meaningful error regions.}
    \label{fig:gp_tsne_0}
    \label{fig:combined_samples}
\end{figure}

\paragraph{Relation to DKL.} Using learned internal representations inside a stationary kernel with learnable hyperparameters is essentially the same as DKL. However, two design choices distinguish REEF-GP from a standard DKL pipeline, and both follow from the post-hoc UQ goal. First, the feature extractor is not trained jointly with the GP: $\mathcal{G}_\theta$ has already produced a physics-aware deformation by solving the PDE under empirical risk minimization, and we reuse that deformation rather than relearning one. The internal layers deform the geometry into a manifold where the physical dynamics are smoothly resolvable, but the resulting embeddings are not, on their own, adapted to provide UQ.

Second, scaling DKL to operator learning datasets ($MN \sim 10^6$) typically relies on sparse approximations such as inducing points \cite{pmlr-v5-titsias09a, pmlr-v38-hensman15}, which impose a low-rank covariance structure that can constrain calibration in ways that are difficult to control without additional regularization. We take a different route, again inspired by KF: rather than committing to a fixed low-rank structure, we condition on random subsets of the training data at every step (Appendix~\ref{appendix:training_algorithm}). KF itself was designed for interpolation and explicitly discards UQ information, so we keep its stochastic-subset philosophy while replacing the RKHS-stability loss with a mini-batch negative log marginal likelihood that retains the calibration signal.

\paragraph{Adapting the Flow for UQ.}

The Transolver embeddings are shaped by predictive accuracy, not by the geometry a stationary kernel would prefer for UQ. We therefore need to compress them into a new latent space whose role is to calibrate uncertainty rather than to relearn the flow. Figure~\ref{fig:gp_tsne_0} illustrates the mechanism. Through the layer selection algorithm we choose three internal representations that are maximally distinct, project them through a spectral-normalized MLP, and use the resulting embedding to augment the input features before feeding them to the stationary kernel. The projection yields a geometry-aware embedding suited for UQ: as seen in Figure~\ref{fig:gp_tsne_0}, the representation tears off at the shock wave regions, bringing some interpretability to the metric space used to quantify uncertainty.

\section{Ablation Studies}\label{appendix:ablation_studies}

We use the \emph{Airfoil} benchmark for the ablation studies. This dataset is sufficiently challenging to surface meaningful differences between design choices, while remaining feasible to retrain across many configurations. The training details and hardware used are described in Appendix \ref{appendix:training_algorithm}.

\subsection{Layer Selection}\label{appendix:ablation:layer_selection}
\begin{figure}[!t]
    \centering
        \includegraphics[width=\textwidth]{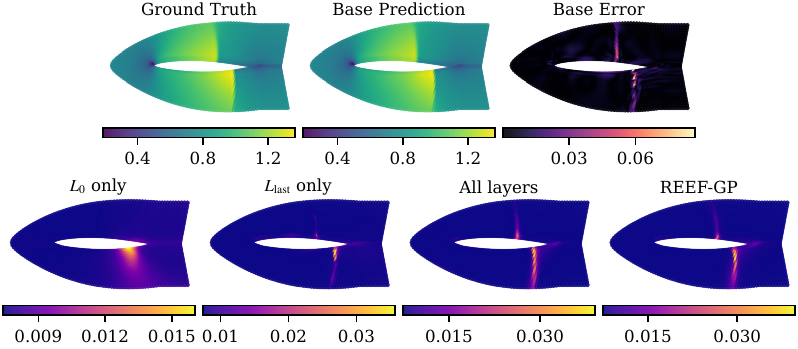}
    \caption{\textbf{Layer-selection ablation on \emph{Airfoil}.} 
    Top: ground truth field, base prediction, absolute error. 
    Bottom: predicted standard deviation under four layer-selection variants. 
    All variants concentrate uncertainty around the shock front; the CKA-selected three-layer combination (REEF-GP) closely matches the full-stack baseline at reduced time and memory costs.}
    \label{fig:ablation_layer_selection_visual}
\end{figure}

We compare four strategies for constructing the augmented state $\mathbf{h}$: (i) using only the input layer $L_0$, (ii) only the output layer $L_{\text{last}}$, (iii) concatenating the full stack of all layers, and (iv) our default CKA-based selection of three layers $\{L_0, l^*, L_{\text{last}}\}$. Results are reported in Table~\ref{tab:ablation_layer_selection} and visualized in Figure~\ref{fig:ablation_layer_selection_visual}. Single-layer variants ($L_0$ or $L_{\text{last}}$ alone) underperform across all probabilistic metrics, with $L_0$ alone being the weakest. Our CKA-selected three-layer combination is statistically indistinguishable from concatenating the full stack (overlapping error bars on every metric), at lower memory and compute cost.

\begin{table}[!b]
\centering
\caption{\textbf{Layer selection ablation.} Effect of the layer set used to construct the augmented state $\mathbf{h}$. Lower is better ($\downarrow$). $L_0$ uses only the input layer; $L_{\text{last}}$ only the output; All layers concatenates the entire stack; \textbf{REEF-GP} selects $\{L_0, l^*, L_{\text{last}}\}$ via CKA. Best per metric in \textbf{bold}, second best \underline{underlined}.}
\label{tab:ablation_layer_selection}
\setlength{\tabcolsep}{4pt}
\renewcommand{\arraystretch}{1.1}
\small
\begin{tabular}{l c c c c c}
\toprule
\textbf{Method} & \textbf{rL2} $\downarrow$ & \textbf{NLL} $\downarrow$ & \textbf{CRPS} $\downarrow$ & \textbf{NIS} $\downarrow$ & \textbf{ES} $\downarrow$ \\
\midrule
\emph{Airfoil (2D)} & (\%) &  & ($\times 10^{-4}$) & ($\times 10^{-3}$) &  \\
\midrule
\textbf{REEF-GP} & \valpmbf{1.24}{0.12} & \valpmul{-3.51}{0.06} & \valpmul{39.80}{2.35} & \valpmul{56.89}{4.66} & \valpmbf{0.45}{0.05} \\
\addlinespace[2pt]
$L_{\text{last}}$ only & \valpmbf{1.24}{0.12} & \valpm{-3.36}{0.12} & \valpm{40.25}{2.41} & \valpm{60.70}{5.10} & \valpmul{0.46}{0.05} \\
$L_0$ only & \valpmbf{1.24}{0.12} & \valpm{-2.75}{0.14} & \valpm{40.89}{2.46} & \valpm{67.72}{5.50} & \valpmul{0.46}{0.05} \\
All layers ($L_0 \dots L_{\text{last}}$) & \valpmbf{1.24}{0.12} & \valpmbf{-3.54}{0.06} & \valpmbf{39.71}{2.40} & \valpmbf{55.99}{4.74} & \valpmbf{0.45}{0.05} \\
\bottomrule
\end{tabular}
\end{table}

\subsection{Number of Training Samples}\label{appendix:ablation:training_samples}

\begin{figure}[!t]
    \centering
        \includegraphics[width=\textwidth]{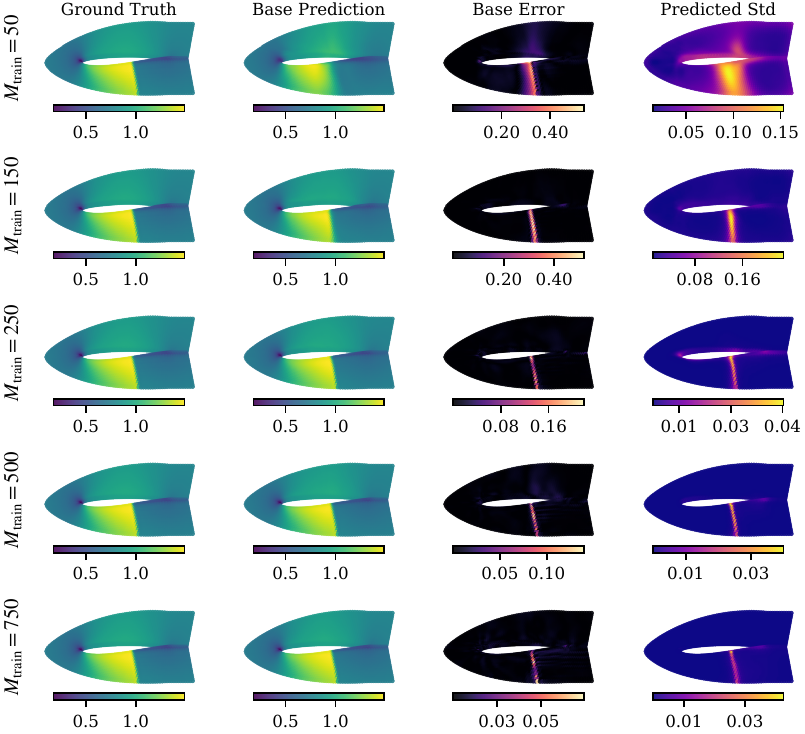}
    \caption{\textbf{Training samples ablation on \emph{Airfoil}.} 
    Each row corresponds to a different training-set size $M_{\text{train}}$, applied jointly to the base Transolver and the GP discrepancy model. 
    Columns: ground truth field, base prediction, absolute error of the base prediction, and predicted standard deviation. 
    As $M_{\text{train}}$ decreases, the base operator's error grows; REEF-GP correctly inflates its predicted uncertainty to track this growth, with the spatial structure of the predicted standard deviation closely matching the spatial structure of the error.}
    \label{fig:ablation_num_train_samples_visual}
\end{figure}

\begin{figure}[!b]
    \centering
        \includegraphics[width=0.4\textwidth]{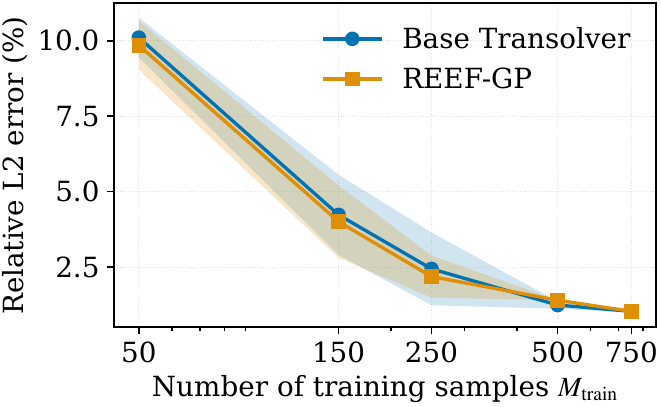}
    \caption{\textbf{Relative L2 error vs.\ training-set size on \emph{Airfoil}.} 
    Both curves show mean across 5 seeds with shaded $\pm 1$ standard deviation. 
    REEF-GP closely follows the base Transolver's accuracy across two orders of magnitude in $M_{\text{train}}$, confirming that adding GP-based uncertainty quantification does not degrade predictive performance.}
    \label{fig:rl2_vs_samples}
\end{figure}

We jointly vary the number of training samples used to train the base Transolver and learn the GP discrepancy model. The test set is held fixed across all configurations. As the operator is trained on more data, its predictions become more accurate, leaving smaller residuals for the GP to model. We observe that REEF-GP adapts gracefully to base operators of varying quality, recovering well-calibrated uncertainty in both data-rich and data-poor regimes. Results are reported in Table~\ref{tab:ablation_num_train_samples} and visualized in Figure~\ref{fig:ablation_num_train_samples_visual} and Figure~\ref{fig:rl2_vs_samples}.

\begin{table}[h]
\centering
\caption{\textbf{Training samples ablation on \emph{Airfoil}.} Effect of the number of training samples used to fit our GP discrepancy model. Lower is better ($\downarrow$). Best per metric in \textbf{bold}, second best \underline{underlined}.}
\label{tab:ablation_num_train_samples}
\setlength{\tabcolsep}{4pt}
\renewcommand{\arraystretch}{1.1}
\small
\begin{tabular}{l c c c c c}
\toprule
\textbf{Method} & \textbf{rL2} $\downarrow$ & \textbf{NLL} $\downarrow$ & \textbf{CRPS} $\downarrow$ & \textbf{NIS} $\downarrow$ & \textbf{ES} $\downarrow$ \\
\midrule
\emph{Airfoil (2D)} & (\%) &  & ($\times 10^{-3}$) & ($\times 10^{-2}$) &  \\
\midrule
$M_{\text{train}} = 50$ & \valpm{9.68}{0.80} & \valpm{0.34}{1.87} & \valpm{32.55}{3.04} & \valpm{62.16}{11.72} & \valpm{3.70}{0.38} \\
\addlinespace[2pt]
$M_{\text{train}} = 150$ & \valpm{3.92}{1.14} & \valpm{-2.17}{0.63} & \valpm{11.22}{2.75} & \valpm{18.45}{5.47} & \valpm{1.49}{0.42} \\
$M_{\text{train}} = 250$ & \valpm{2.18}{0.68} & \valpm{-3.07}{0.06} & \valpm{6.35}{1.46} & \valpm{9.11}{0.87} & \valpm{0.80}{0.22} \\
\textbf{$M_{\text{train}} = 500$} & \valpmul{1.24}{0.12} & \valpmul{-3.51}{0.06} & \valpmul{3.98}{0.24} & \valpmul{5.69}{0.47} & \valpmul{0.45}{0.05} \\
$M_{\text{train}} = 750$ & \valpmbf{1.03}{0.03} & \valpmbf{-3.60}{0.02} & \valpmbf{3.46}{0.04} & \valpmbf{5.06}{0.08} & \valpmbf{0.38}{0.01} \\
\bottomrule
\end{tabular}
\end{table}

\subsection{Stochastic Training Subset Size}\label{appendix:ablation:subset_size}

\begin{figure}[h]
    \centering
        \includegraphics[width=\textwidth]{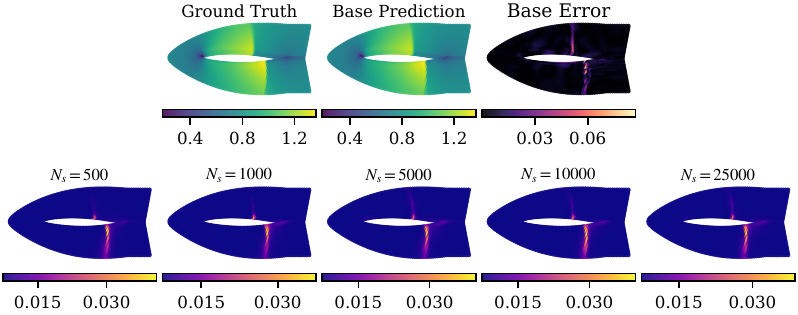}
    \caption{\textbf{Subset-size ablation on \emph{Airfoil}.} 
    Top: ground truth field, base prediction, absolute error. 
    Bottom: predicted standard deviation under five subset sizes $N_s$. 
    All variants concentrate uncertainty around the shock front.}
    \label{fig:ablation_subset_size_visual}
\end{figure}

We ablate the per-expert support-set size $N_s$ used both during stochastic training (each gradient step conditions the GP on a random size-$N_s$ subset of the training residuals) and during inference (each gPoE expert is conditioned on an independent size-$N_s$ subset). Larger $N_s$ provides each GP with more conditioning data per step, at the cost of $\mathcal{O}(N_s^3)$ per Cholesky factorization. Results are reported in Table~\ref{tab:ablation_subset_size} and visualized in Figure~\ref{fig:ablation_subset_size_visual}.
We observe that REEF-GP recovers the same calibration quality across two orders of magnitude in $N_s$ (from $500$ to $25{,}000$). Figure~\ref{fig:ablation_subset_size_visual} confirms this visually for a random test sample: the predicted uncertainty fields are nearly identical across all five subset sizes, concentrating around the shock front in every variant. Following these observations, we adopt $N_s = 5000$ as a middle-of-the-range default that balances conditioning quality against the cubic cost of Cholesky factorization.

\begin{table}[!t]
\centering
\caption{\textbf{Subset-size ablation on \emph{Airfoil}.} Effect of the per-expert support-set size $N_s$ used at inference. For the results in text we use $N_s = 5000$. Lower is better ($\downarrow$).} 
\label{tab:ablation_subset_size}
\setlength{\tabcolsep}{4pt}
\renewcommand{\arraystretch}{1.1}
\small
\begin{tabular}{l c c c c c}
\toprule
\textbf{Method} & \textbf{rL2} $\downarrow$ & \textbf{NLL} $\downarrow$ & \textbf{CRPS} $\downarrow$ & \textbf{NIS} $\downarrow$ & \textbf{ES} $\downarrow$ \\
\midrule
\emph{Airfoil (2D)} & (\%) &  & ($\times 10^{-4}$) & ($\times 10^{-3}$) &  \\
\midrule
$N_s = 500$ & \valpm{1.24}{0.12} & \valpm{-3.50}{0.08} & \valpm{39.91}{2.39} & \valpm{57.00}{5.53} & \valpm{0.45}{0.05} \\
\addlinespace[2pt]
$N_s = 1000$ & \valpm{1.24}{0.12} & \valpm{-3.50}{0.08} & \valpm{39.87}{2.41} & \valpm{56.81}{5.92} & \valpm{0.45}{0.05} \\
$N_s = 5000$ & \valpm{1.24}{0.12} & \valpm{-3.51}{0.06} & \valpm{39.80}{2.35} & \valpm{56.89}{4.66} & \valpm{0.45}{0.05} \\
$N_s = 10000$ & \valpm{1.24}{0.12} & \valpm{-3.51}{0.06} & \valpm{39.80}{2.36} & \valpm{56.94}{4.64} & \valpm{0.45}{0.05} \\
$N_s = 25000$ & \valpm{1.24}{0.12} & \valpm{-3.51}{0.06} & \valpm{39.80}{2.38} & \valpm{57.30}{4.68} & \valpm{0.45}{0.05} \\
\bottomrule
\end{tabular}
\end{table}

\subsection{Number of Experts}\label{appendix:ablation:number_experts}

\begin{figure}[!t]
    \centering
        \includegraphics[width=\textwidth]{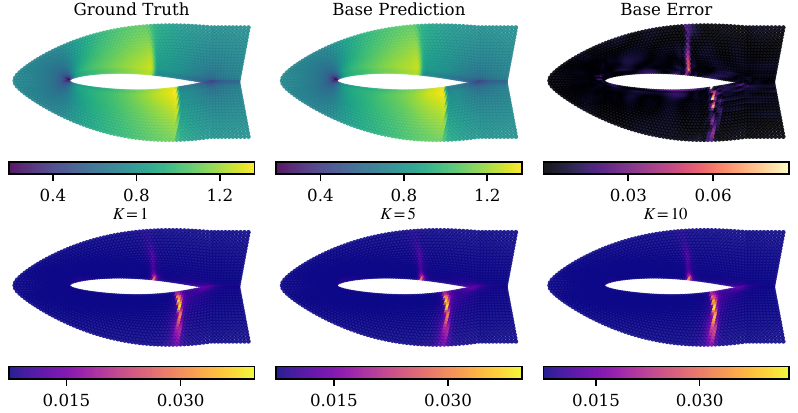}
    \caption{\textbf{Number of gPoE experts ablation on \emph{Airfoil}.} 
    Top: ground truth field, base prediction, absolute error. 
    Bottom: predicted standard deviation under three values of $K$, 
    the number of GP experts in the gPoE inference scheme.}
    \label{fig:ablation_num_experts_visual}
\end{figure}

We compare three values of $K$, the number of GP experts used in the 
gPoE inference scheme: $K = 1$, our default $K = 5$, and $K = 10$. 
Results on \emph{Airfoil} are reported in 
Table~\ref{tab:ablation_num_experts} and visualized in 
Figure~\ref{fig:ablation_num_experts_visual}. Calibration is 
essentially insensitive to $K$ on this benchmark: all three settings 
produce statistically indistinguishable metrics.

\begin{table}[!b]
\centering
\caption{\textbf{Number of gPoE experts ablation on \emph{Airfoil}.} 
Effect of the number of GP experts $K$ used in the generalized Product 
of Experts inference scheme. Lower is better ($\downarrow$).}
\label{tab:ablation_num_experts}
\setlength{\tabcolsep}{4pt}
\renewcommand{\arraystretch}{1.1}
\small
\begin{tabular}{l c c c c c}
\toprule
\textbf{Method} & \textbf{rL2} $\downarrow$ & \textbf{NLL} $\downarrow$ & \textbf{CRPS} $\downarrow$ & \textbf{NIS} $\downarrow$ & \textbf{ES} $\downarrow$ \\
\midrule
\emph{Airfoil (2D)} & (\%) &  & ($\times 10^{-4}$) & ($\times 10^{-3}$) &  \\
\midrule
$K=1$ & \valpm{1.24}{0.12} & \valpm{-3.51}{0.06} & \valpm{39.82}{2.36} & \valpm{56.90}{4.65} & \valpm{0.45}{0.05} \\
\addlinespace[2pt]
$K=5$ & \valpm{1.24}{0.12} & \valpm{-3.51}{0.06} & \valpm{39.80}{2.35} & \valpm{56.89}{4.66} & \valpm{0.45}{0.05} \\
$K=10$ & \valpm{1.24}{0.12} & \valpm{-3.51}{0.06} & \valpm{39.80}{2.35} & \valpm{56.89}{4.66} & \valpm{0.45}{0.05} \\
\bottomrule
\end{tabular}
\end{table}

\subsection{Heteroscedastic vs. Homoscedastic Noise}\label{appendix:ablation:noise}

\begin{figure}[!t]
    \centering
        \includegraphics[width=\textwidth]{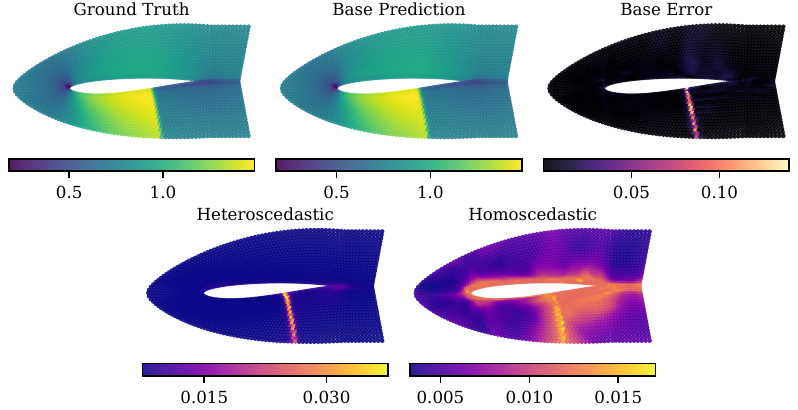}
    \caption{\textbf{Heteroscedastic vs. homoscedastic noise ablation on \emph{Airfoil}.} 
    Top: ground truth, base prediction, absolute error. 
    Bottom: predicted standard deviation under each noise model. 
    The heteroscedastic model concentrates uncertainty around the shock front 
    where the operator's error is largest.}
    \label{fig:ablation_noise_visual}
\end{figure}

We compare our heteroscedastic noise model $\sigma^2_{\psib}(\augz)$, which is input-dependent and parameterized by an MLP, against a homoscedastic baseline that fits a single scalar noise variance. Results are reported in Table~\ref{tab:ablation_noise} and visualized in Figure~\ref{fig:ablation_noise_visual}. The two noise models yield comparable aggregate metrics, with overlapping error bars on every probabilistic score. The qualitative difference is spatial: both models concentrate uncertainty around the shock front, but the heteroscedastic model produces a sharper, higher-magnitude uncertainty signal localized to the error region, while the homoscedastic model assigns non-negligible uncertainty across regions where the operator's error is small, with a compressed numerical range overall. We adopt the heteroscedastic model as our default as it provides more structurally meaningful uncertainties.

\begin{table}[h]
\centering
\caption{\textbf{Heteroscedastic vs. homoscedastic noise ablation on \emph{Airfoil}.} Effect of the noise model used in the GP likelihood. Heteroscedastic uses an input-dependent variance $\sigma^2_{\psib}(\augz)$; homoscedastic uses a single scalar variance.} 
\label{tab:ablation_noise}
\setlength{\tabcolsep}{4pt}
\renewcommand{\arraystretch}{1.1}
\small
\begin{tabular}{l c c c c c}
\toprule
\textbf{Method} & \textbf{rL2} $\downarrow$ & \textbf{NLL} $\downarrow$ & \textbf{CRPS} $\downarrow$ & \textbf{NIS} $\downarrow$ & \textbf{ES} $\downarrow$ \\
\midrule
\emph{Airfoil (2D)} & (\%) &  & ($\times 10^{-4}$) & ($\times 10^{-3}$) &  \\
\midrule
Heteroscedastic & \valpm{1.24}{0.12} & \valpm{-3.51}{0.06} & \valpm{39.80}{2.35} & \valpm{56.89}{4.66} & \valpm{0.45}{0.05} \\
\addlinespace[2pt]
Homoscedastic & \valpm{1.24}{0.12} & \valpm{-3.41}{0.14} & \valpm{38.82}{2.66} & \valpm{58.84}{5.84} & \valpm{0.45}{0.05} \\
\bottomrule
\end{tabular}
\end{table}

\clearpage

\section{Base Neural Operator Configuration}\label{appendix:base_no}

To ensure a rigorous evaluation of our post-hoc UQ framework, we train a deterministic Transolver \cite{wu2024transolverfasttransformersolver} model as required for each of the benchmark tasks and baseline configurations. The architecture is configured based on the original setup to achieve near state-of-the-art predictive accuracy.

\paragraph{Transolver.}
The model architecture consists of a sequence of tranformer blocks that rely on a Physics-Attention mechanism that adaptively groups the discretized spatial domain into learnable slices. Table \ref{tab:transolver_hyperparams} details the specific configurations used for the evaluated benchmarks. Across all tasks, we use the GELU \cite{hendrycks2023gaussianerrorlinearunits} activation function and optimize the network using AdamW \cite{loshchilov2019decoupledweightdecayregularization}. The learning rate is decayed using a OneCycle learning rate scheduler to ensure stable convergence. We use the relative $L^2$ error as the loss function. All models are trained for a fixed number of epochs, with results averaged over five random seeds to ensure statistical robustness.

\begin{table}[hbtp]
    \centering
    \caption{\textbf{Architecture and hyperparameter configurations.} The base Transolver slightly varies across the benchmarks.}
    \label{tab:transolver_hyperparams}
    \begin{tabular}{lccccc}
        \toprule
        \textbf{Hyperparameter} & \textbf{Elasticity} & \textbf{Airfoil}  & \textbf{Pipe} & \textbf{ShapeNet Car} & \textbf{Ahmed Body} \\
        \midrule
        Hidden channels ($D_h$) & 128 & 128 & 128 & 256 & 256 \\
        Transformer layers ($L$) & 8 & 8 & 8 & 8 & 8 \\
        Physics slices ($S$) & 64 & 64 & 64 & 32 & 32 \\
        Attention heads ($H$) & 8 & 8 & 8 & 8 & 8 \\
        \midrule
        Epochs & 500 & 500 & 500 & 500 & 500 \\
        Batch size & 1 & 4 & 4 & 1 & 1 \\
        Initial learning rate & $1 \times 10^{-3}$ & $1 \times 10^{-3}$ & $1 \times 10^{-3}$ & $1 \times 10^{-3}$ & $1 \times 10^{-3}$ \\
        Weight decay & $1 \times 10^{-4}$ & $1 \times 10^{-4}$ & $1 \times 10^{-4}$ & $1 \times 10^{-4}$ & $1 \times 10^{-4}$ \\
        \bottomrule
    \end{tabular}
\end{table}

\paragraph{Dataset Splits for UQ Evaluation.}
While standard operator learning benchmarks often assume abundant data, the value of UQ is most pronounced in data-constrained environments where the surrogate model is forced to generalize. To rigorously test the calibration of REEF-GP and the baselines under these conditions, we evaluate all main experiments using a reduced data split. Table \ref{tab:transolver_data_splits} shows the data splits for each dataset.

\begin{table}[hbtp]
    \centering
    \caption{\textbf{Dataset splits.} The number of samples slightly varies across the five benchmarks.}
    \label{tab:transolver_data_splits}
    \begin{tabular}{lccccc}
        \toprule
        \textbf{} & \textbf{Elasticity} & \textbf{Airfoil} & \textbf{Pipe} & \textbf{ShapeNet Car} & \textbf{Ahmed Body} \\
        \midrule
        Train ($M_{train}$) & 500 & 500 & 500 & 500 & 400\\
        Validation ($M_{val}$) & 100 & 100 & 100 & 100 & 50\\
        Test ($M_{test}$) & 250 & 250 & 250 & 250 & 100\\
        \bottomrule
    \end{tabular}
\end{table}

\section{Benchmarks}\label{appendix:benchmarks}

We use three 2D datasets and two 3D datasets to evaluate REEF-GP against competing methods detailed in Appendix \ref{appendix:uq_baselines}.

\paragraph{Elasticity.} This benchmark provides the stress field of an elastic material subject to a tension traction $\mathbf{t}$ applied on the top edge of a unit cell with an irregular shape void in the interior \cite{10.5555/3648699.3649087}. There are a total of $M=2000$ samples. For each sample the input domain is discretized as a point cloud comprising $N = 972$ spatial nodes.

\paragraph{Airfoil.} This benchmark provides the resulting Mach number field from a transonic flow over 2D airfoil geometries, governed by the Euler equation \cite{10.5555/3648699.3649087}. The spatial domain is discretized into a structured mesh around the airfoil which we fix to $N = 2820$ points per sample. There are a total of $M = 2490$ samples.

\paragraph{Pipe.} This benchmark provides the fluid velocity through a pipe of changing shape \cite{10.5555/3648699.3649087}. The input domain is discretized via $N = 129\times 27$ spatial points and there are a total of $M=2310$ samples.

\paragraph{ShapeNet Car.} This benchmark provides the surface pressure field resulting from turbulent aerodynamic flow over diverse 3D vehicle geometries extracted from the "car" category of ShapeNet \cite{chang2015shapenetinformationrich3dmodel}, governed by the Navier-Stokes equations \cite{10.1145/3197517.3201325}. The input domain for each vehicle surface is discretized as an unstructured point cloud, which we fix to $N=3500$ spatial points per sample. There are a total of $M=889$ samples.

\paragraph{Ahmed Body.} This benchmark provides vehicle aerodynamics simulations based on the Ahmed-body shapes \cite{li2023geometryinformed}. There are a total of $M=551$ shapes and we fix the spatial points to $N=5000$ per sample. As additional inputs we consider the inlet velocity and Reynolds number as they change across samples.

\section{UQ Baselines Configuration}\label{appendix:uq_baselines}

To assess the performance of our post-hoc UQ framework, we compare it against six established UQ methodologies. To ensure a fair and rigorous comparison, all baselines are implemented using the exact same base Transolver architecture and evaluated over the same dataset splits.

\paragraph{Deep Ensembles.} 
We ensemble $5$ independent Transolver models, each of them trained from scratch using the same hyperparameters but initialized with a different random seed. The final predictive mean and variance are computed empirically from the ensemble's $5$ distinct forward passes. While highly accurate, this method incurs a $5\times$ multiplier in total training cost.

\paragraph{Monte Carlo (MC) Dropout.} 
 Following standard practices, dropout is injected directly into the transformer blocks. Specifically, within the Physics-Attention mechanism, dropout is applied in two distinct locations: first, directly to the attention matrix after the softmax operation (randomly dropping query-key connections prior to multiplication with the value matrix), and second, to the output of the linear projection immediately following the deslicing step, before the tensor is passed to subsequent network components. We use a standard dropout rate of $p=0.1$. At inference time, dropout is kept active, and we draw $50$ stochastic forward passes to compute the predictive mean and variance.

\paragraph{Probabilistic Neural Operator (PNO).} 
PNO \cite{blte2025probabilistic} extends the MC Dropout approach by optimizing the network directly for probabilistic evaluation during training. Using Transolver as the base architecture with the identical dropout locations described above, we implement PNO by modifying the training objective. Instead of the deterministic relative $L^2$ error, PNO minimizes the empirical Energy Score (ES) defined in Equation \ref{eq:energy_score} on the training set. To balance computational feasibility with stable training, we use $K=5$ forward passes during training, and $K=50$ forward passes during evaluation.

\paragraph{Function-Valued Laplace Approximation (LUNO-LA).} 
We implement a last-layer Laplace approximation derived from the LUNO framework \cite{magnani2025linearization} using the \texttt{laplace-torch} library \cite{NEURIPS2021_a7c95857}. LUNO leverages the concept of \textit{currying} to translate weight-space uncertainty into a continuous, function-valued Gaussian Process over the spatial domain. Because Transolver relies on standard pointwise linear projections in its final layers (rather than complex spectral convolutions), the network mapping is inherently linear with respect to the last layer's weights. Thus, applying a Laplace approximation to the final layer natively yields the exact LUNO covariance kernel $k(x, x') = \mathbf{h}(x)^\top \Sigma \mathbf{h}(x')$, where $\mathbf{h}(x)$ are the spatially-aligned Transolver features and $\Sigma$ is the weight posterior covariance computed using a full Hessian structure. In our experiments we find that utilizing all spatial points makes covariance construction and inversion memory-prohibitive and unstable, preventing convergence to a robust solution. To overcome this bottleneck and ensure well-conditioned covariance inversion, we uniformly subsample a maximum of $50k$ spatial points across the dataset during optimization. The prior precision of the Laplace approximation is optimized via grid search on a validation set.

\paragraph{Input Perturbation.} 
Drawing on test-time data augmentation \cite{ayhan2018testtime} and initial-condition perturbation \cite{pathak2022fourcastnetglobaldatadrivenhighresolution}, we derive predictive uncertainty from a fully trained, deterministic Transolver model's sensitivity to input variations. At inference time, we inject independent and identically distributed (i.i.d.) Gaussian noise ($\sigma_n = 10^{-2}$) directly into the input tensors. For each test sample, we execute $50$ independent forward passes, sampling a new noise mask for the inputs prior to each pass. The final predictive mean and variance are computed empirically across these $50$ stochastic forward passes.

\paragraph{Deep Vecchia Ensemble (DVE-spatial).} 
We adapt the Deep Vecchia Ensemble (DVE) \cite{pmlr-v258-jimenez25a}, utilizing the author's original implementation, which constructs an ensemble of GPs over the internal hidden-layer representations of a trained neural network. The original DVE computes the nearest-neighbor conditioning sets for its Vecchia approximation using distances strictly in the intermediate feature space. However, in dense spatial and operator learning tasks, geometrically neighboring points frequently exhibit nearly identical embeddings, which leads to severely ill-conditioned covariance matrices. To evaluate DVE as a viable baseline in our setting, we adapt it into \emph{DVE-spatial} by concatenating the physical spatial coordinates to the intermediate embeddings prior to computing the distance metrics.  Although DVE was designed for standard regression rather than operator learning and it might require further specialized modifications for this domain, it serves as a highly relevant benchmark for UQ applied directly within a network's feature space. Because the layer-wise nearest-neighbor searches and GP inference are computationally expensive, we configure DVE-spatial to balance its cost based on its default settings. Specifically, we restrict the Vecchia nearest-neighbor conditioning set size to $m=8$ and utilize an approximate nearest neighbor search by partitioning the feature space into $500$ cells and searching across the $15$ closest cells for each query.

\section{Metrics}\label{appendix:metrics}

We evaluate predictive accuracy with the relative $L^2$ error, and probabilistic calibration with four standard scoring rules: NLL, CRPS, NIS, and ES. Throughout, $M_{\text{test}}$ denotes the number of test instances; for instance $i$, $\outv_i \in \mathbb{R}^N$ is the ground-truth nodal solution with entries $v_{i,j}$, and the predictive distribution at node $j$ is Gaussian with mean $\mu_{i,j}$ and variance $\sigma^2_{i,j}$.

\paragraph{Relative $L^2$ Error.}
Quantifies the predictive accuracy of the operator:
\begin{equation}
    \mathrm{rL2} = \frac{1}{M_{\text{test}}} \sum_{i=1}^{M_{\text{test}}} \frac{\|\trueop(u_i, a_i) - \paramnop(\inu_i, \ingeom_i)\|_{\mathcal{V}}}{\|\trueop(u_i, a_i)\|_{\mathcal{V}}},
\end{equation}
where $\|\cdot\|_{\mathcal{V}}$ is the $L^2$ norm over the physical domain.

\paragraph{Negative Log-Likelihood (NLL).}
A strictly proper scoring rule that penalizes both miscalibration and overconfidence:
\begin{equation}
    \mathrm{NLL} = \frac{1}{M_{\text{test}} N} \sum_{i=1}^{M_{\text{test}}} \sum_{j=1}^{N} \left[ \frac{1}{2} \log(2\pi \sigma^2_{i,j}) + \frac{(v_{i,j} - \mu_{i,j})^2}{2\sigma^2_{i,j}} \right].
\end{equation}

\paragraph{Continuous Ranked Probability Score (CRPS).}
A probabilistic generalization of the mean absolute error. For Gaussian predictives it admits the closed form:
\begin{equation}
    \mathrm{CRPS} = \frac{1}{M_{\text{test}} N} \sum_{i=1}^{M_{\text{test}}} \sum_{j=1}^{N} \sigma_{i,j}\!\left[ z_{i,j}\bigl(2\Phi(z_{i,j}) - 1\bigr) + 2\phi(z_{i,j}) - \tfrac{1}{\sqrt{\pi}} \right],
\end{equation}
with standardized error $z_{i,j} = (v_{i,j} - \mu_{i,j})/\sigma_{i,j}$, and $\Phi, \phi$ the standard-normal CDF and PDF.

\paragraph{Negatively Oriented Interval Score (NIS).}
Evaluates the quality of central prediction intervals. For confidence level $1-\alpha$, let $L_{i,j}$ and $U_{i,j}$ be the predicted lower and upper bounds at node $j$:
\begin{equation}
    \mathrm{NIS} = \frac{1}{M_{\text{test}} N} \sum_{i=1}^{M_{\text{test}}} \sum_{j=1}^{N} \left[ (U_{i,j} - L_{i,j}) + \frac{2}{\alpha}(L_{i,j} - v_{i,j})_+ + \frac{2}{\alpha}(v_{i,j} - U_{i,j})_+ \right],
\end{equation}
where $(x)_+ = \max(x, 0)$. The first term rewards sharpness; the latter two penalize miscoverage. We report NIS at $\alpha = 0.05$.

\paragraph{Energy Score (ES).}
A multivariate proper scoring rule, estimated by Monte Carlo with $K = 50$ samples $\outv_i^{(k)}$ drawn from the predictive:
\begin{equation}
    \mathrm{ES} = \frac{1}{M_{\text{test}}} \sum_{i=1}^{M_{\text{test}}} \left[ \frac{1}{K} \sum_{k=1}^{K} \|\outv_i^{(k)} - \outv_i\|_2 - \frac{1}{2K(K-1)} \sum_{k \neq k'} \|\outv_i^{(k)} - \outv_i^{(k')}\|_2 \right].
    \label{eq:energy_score}
\end{equation}
The first term rewards accuracy; the second penalizes overconfidence by rewarding sample diversity.

\section{Additional Results}\label{appendix:additional_results}

\begin{figure}[b] 
    \centering 
        \includegraphics[width=\linewidth]{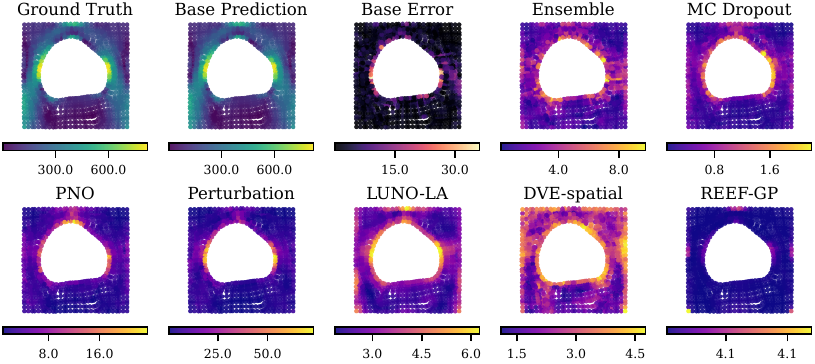} 
    \caption{\textbf{Predictive standard deviation fields on \emph{Elasticity}.}  The rows show in order: ground truth, base prediction, base error and the standard deviation of a sample for each of the UQ baselines.}
    \label{fig:elasticity_main} 
\end{figure}

\begin{figure}[t] 
    \centering 
        \includegraphics[width=\linewidth]{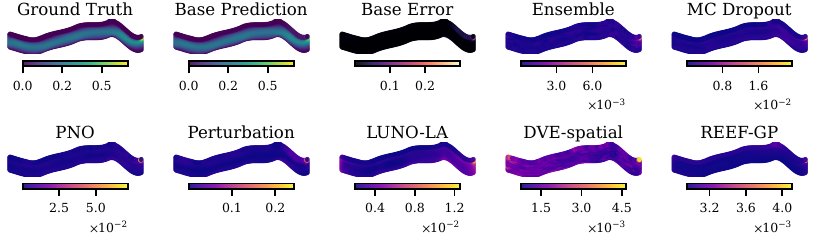} 
    \caption{\textbf{Predictive standard deviation fields on \emph{Pipe}.}  The rows show in order: ground truth, base prediction, base error and the standard deviation of a sample for each of the UQ baselines.}
    \label{fig:pipe_main} 
\end{figure}

\begin{figure}[t] 
    \centering 
        \includegraphics[width=\linewidth]{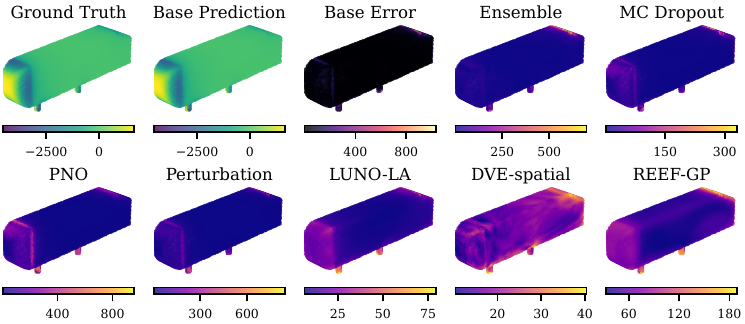} 
    \caption{\textbf{Predictive standard deviation fields on \emph{Ahmed}.}  The rows show in order: ground truth, base prediction, base error and the standard deviation of a sample for each of the UQ baselines.} 
    \label{fig:ahmed_main} 
\end{figure}

\begin{figure}[h]
    \centering
        \includegraphics[width=\textwidth]{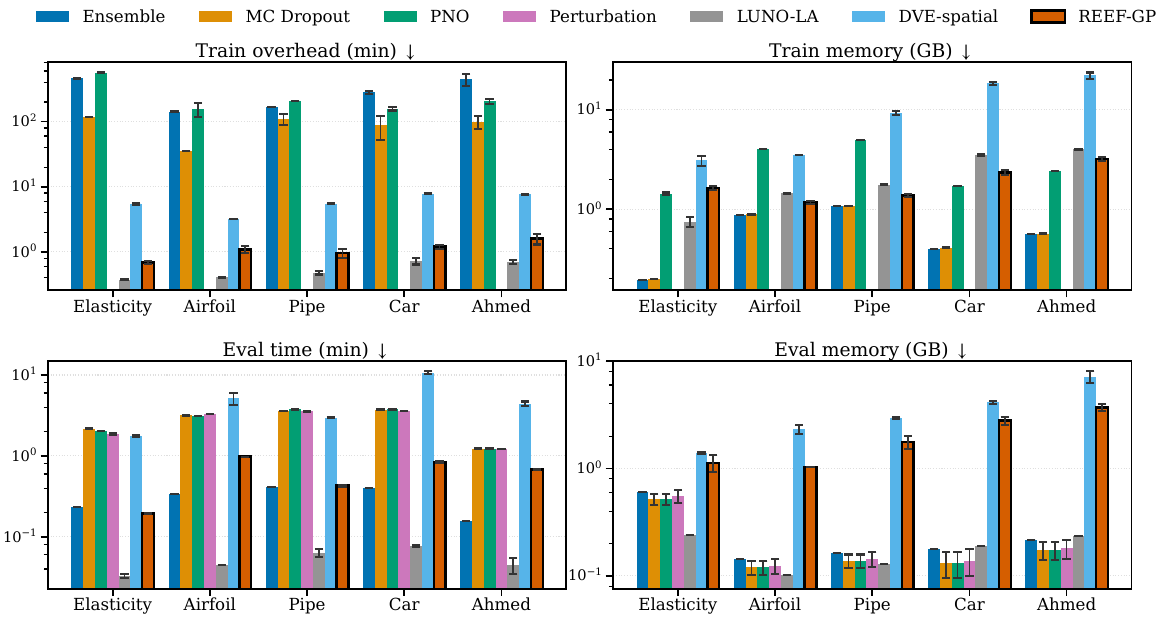}
    \caption{\textbf{Compute costs on all benchmarks:} Train overhead (the additional time needed on top of a trained base operator to enable UQ), evaluation time, peak train memory, and peak evaluation memory on all benchmarks. REEF-GP maintains an excellent balance between training overhead and inference speed, while peak evaluation memory is higher.}
    \label{fig:compute_cost_appendix}
\end{figure}

This section presents the results of REEF-GP and the UQ baselines on the remaining three benchmarks: \emph{Elasticity} (2D), \emph{Pipe} (2D), and \emph{Ahmed Body} (3D). The quantitative results are reported in Table~\ref{tab:results_appendix}, while spatial uncertainty visualizations and compute cost trade-offs are provided in Figures~\ref{fig:elasticity_main}--\ref{fig:ahmed_main} and Figure~\ref{fig:compute_cost_appendix}, respectively.

\paragraph{Consistent preservation of predictive accuracy.} REEF-GP successfully preserves the predictive accuracy of the base Transolver across all three additional datasets. As shown in Table~\ref{tab:results_appendix}, the rL2 of REEF-GP matches that of the deterministic Base model perfectly. In contrast, train-time modifications such as MC Dropout, PNO, and DVE-spatial consistently suffer from degraded rL2 performance.

\paragraph{Competitive probabilistic performance.}
REEF-GP remains highly competitive across all probabilistic metrics. Most notably, on the complex 3D \emph{Ahmed Body} dataset, REEF-GP achieves an NLL of $4.21$, effectively tying with the computationally expensive Deep Ensembles ($4.20$) while outperforming all other post-hoc and train-time baselines. On the 2D benchmarks (\emph{Elasticity} and \emph{Pipe}), REEF-GP yields well-calibrated posteriors with NLL and CRPS scores closely tracking the top-performing methods.

\paragraph{Spatial alignment of uncertainty.}
Figures~\ref{fig:elasticity_main}, \ref{fig:pipe_main}, and \ref{fig:ahmed_main} further confirm the spatial coherence of REEF-GP. The predicted standard deviation tracks the localized prediction errors of the base operator, although in \emph{Elasticity} benchmark our REEF-GP uncertainty is quite regular across the entire domain.

\textbf{Cost-quality tradeoff.} Figure~\ref{fig:compute_cost_appendix} compares the compute cost across all UQ methods on all available benchmarks. It confirms that REEF-GP consistently achieves a training overhead an order of magnitude lower than Deep Ensembles while remaining competitive on evaluation time and train memory. The only tradeoff is in evaluation memory, although it remains well within commodity GPU memory.

\begin{table}[!t]
\centering
\caption{\textbf{Evaluation metrics for the three benchmarks not in the main text.} Lower is better ($\downarrow$). Best in \textbf{bold}, second best \underline{underlined}. Rankings exclude Base. See Appendix \ref{appendix:metrics} for the definition of the metrics.}
\label{tab:results_appendix}
\setlength{\tabcolsep}{4pt}
\renewcommand{\arraystretch}{1.1}
\small
\begin{tabular}{l c c c c c}
\toprule
\textbf{Method} & \textbf{rL2} $\downarrow$ & \textbf{NLL} $\downarrow$ & \textbf{CRPS} $\downarrow$ & \textbf{NIS} $\downarrow$ & \textbf{ES} $\downarrow$ \\
\midrule
\emph{Elasticity (2D)} & (\%) &  &  &  &  \\
\midrule
Base (no UQ) & \valpm{1.68}{0.16} & -- & -- & -- & -- \\
\addlinespace[2pt]
Ensemble & $\mathbf{1.24}$ & $\mathbf{2.45}$ & $\mathbf{1.30}$ & $\mathbf{13.88}$ & $\mathbf{53.73}$ \\
MC Dropout & \valpmul{1.58}{0.12} & \valpm{19.87}{3.22} & \valpm{2.06}{0.14} & \valpm{59.76}{4.89} & \valpm{97.29}{7.60} \\
PNO & \valpm{2.39}{0.14} & \valpm{3.08}{0.07} & \valpm{2.65}{0.15} & \valpm{30.58}{2.04} & \valpm{118.18}{7.12} \\
Perturbation & \valpm{2.17}{0.14} & \valpm{2.85}{0.05} & \valpm{2.43}{0.10} & \valpm{29.13}{0.71} & \valpm{125.12}{4.99} \\
LUNO-LA & \valpm{1.68}{0.16} & \valpmul{2.80}{0.36} & \valpmul{1.85}{0.18} & \valpmul{25.32}{5.82} & \valpm{87.82}{10.50} \\
DVE-spatial & \valpm{5.25}{0.23} & \valpm{10.48}{0.73} & \valpm{6.36}{0.30} & \valpm{162.11}{10.59} & \valpm{317.43}{15.28} \\
\textbf{REEF-GP} & \valpm{1.68}{0.16} & \valpm{2.83}{0.10} & \valpm{1.97}{0.15} & \valpm{25.94}{2.70} & \valpmul{86.37}{7.96} \\
\midrule
\emph{Pipe (2D)} & (\%) &  & ($\times 10^{-4}$) & ($\times 10^{-3}$) & ($\times 10^{-2}$) \\
\midrule
Base (no UQ) & \valpm{0.52}{0.02} & -- & -- & -- & -- \\
\addlinespace[2pt]
Ensemble & $\mathbf{0.42}$ & $\mathbf{-6.32}$ & $\mathbf{2.61}$ & $\mathbf{3.68}$ & $\mathbf{3.38}$ \\
MC Dropout & \valpm{0.64}{0.03} & \valpmul{-5.58}{0.08} & \valpmul{5.12}{0.20} & \valpmul{6.96}{0.28} & \valpmul{5.78}{0.29} \\
PNO & \valpm{0.77}{0.05} & \valpm{-5.12}{0.05} & \valpm{7.77}{0.36} & \valpm{11.07}{0.51} & \valpm{7.68}{0.46} \\
Perturbation & \valpm{2.49}{0.25} & \valpm{-4.22}{0.10} & \valpm{21.18}{1.59} & \valpm{26.35}{0.81} & \valpm{22.28}{1.35} \\
LUNO-LA & \valpmul{0.52}{0.02} & \valpm{-4.63}{0.15} & \valpm{10.11}{1.30} & \valpm{16.61}{2.16} & \valpm{8.63}{0.90} \\
DVE-spatial & \valpm{3.69}{0.89} & \valpm{0.60}{1.60} & \valpm{33.98}{8.33} & \valpm{69.72}{12.12} & \valpm{35.29}{7.82} \\
\textbf{REEF-GP} & \valpmul{0.52}{0.02} & \valpm{-4.73}{0.02} & \valpm{8.19}{0.09} & \valpm{13.85}{0.29} & \valpm{7.21}{0.16} \\
\midrule
\emph{Ahmed Body (3D)} & (\%) &  &  &  & ($\times 10^{3}$) \\
\midrule
Base (no UQ) & \valpm{6.62}{0.05} & -- & -- & -- & -- \\
\addlinespace[2pt]
Ensemble & $\mathbf{5.98}$ & $\mathbf{4.20}$ & $\mathbf{6.27}$ & $\mathbf{104.01}$ & $\mathbf{1.03}$ \\
MC Dropout & \valpm{6.94}{0.42} & \valpm{15.10}{5.25} & \valpm{8.96}{1.42} & \valpm{227.43}{37.79} & \valpm{1.34}{0.07} \\
PNO & \valpm{11.59}{0.94} & \valpm{8.38}{1.68} & \valpm{15.43}{1.63} & \valpm{244.62}{30.07} & \valpm{1.81}{0.15} \\
Perturbation & \valpm{7.51}{0.14} & \valpm{6.88}{0.80} & \valpm{8.43}{0.11} & \valpm{144.79}{3.94} & \valpm{1.24}{0.02} \\
LUNO-LA & \valpmul{6.62}{0.05} & \valpm{16.95}{10.78} & \valpmul{8.07}{0.32} & \valpm{206.69}{50.88} & \valpm{1.34}{0.09} \\
DVE-spatial & \valpm{13.54}{1.59} & \valpm{7.90}{1.72} & \valpm{17.66}{1.52} & \valpm{400.41}{24.53} & \valpm{2.65}{0.24} \\
\textbf{REEF-GP} & \valpm{6.63}{0.05} & \valpmul{4.21}{0.08} & \valpm{8.88}{0.79} & \valpmul{136.68}{3.12} & \valpmul{1.19}{0.01} \\
\bottomrule
\end{tabular}
\end{table}

\clearpage

\section{Geometric OOD}\label{appendix:geometric_OOD}

\paragraph{Maximum Mean Discrepancy (MMD).}

Given two point clouds $\mathcal{X}$ and $\mathcal{Y}$, each comprising $P$ points, the squared MMD is defined as:
\begin{equation}
    MMD^2(\mathcal{X}, \mathcal{Y}) = \frac{1}{P^2} \sum_{i=1}^P \sum_{j=1}^P k(x_i, x_j) + \frac{1}{P^2} \sum_{i=1}^P \sum_{j=1}^P k(y_i, y_j) - \frac{2}{P^2} \sum_{i=1}^P \sum_{j=1}^P k(x_i, y_j)
\end{equation}
where $k(\cdot, \cdot)$ denotes the Gaussian (Radial Basis Function) kernel, given by $k(x,y) = \exp\left(-\frac{||x-y||^2}{2l^2}\right)$. To appropriately scale the kernel to the data distribution, the lengthscale parameter $l$ is determined via a median heuristic computed as the median of the pairwise $L_2$ distances sampled from the training point clouds.

\begin{table}[!b]
\centering
\caption{\textbf{OOD robustness across MMD-based geometric quantiles.} Test samples are partitioned into four groups by MMD distance from the training distribution: Q1 (closest, in-distribution) through Q4 (furthest, most geometrically novel). For each method we report rL2 ($\downarrow$) and NLL ($\downarrow$); rL2 in percent. Best in \textbf{bold}, second best \underline{underlined}.}
\label{tab:results_quantile_ood}
\setlength{\tabcolsep}{3pt}
\renewcommand{\arraystretch}{1.05}
\footnotesize
\resizebox{\linewidth}{!}{%
\begin{tabular}{l c c c c c c c c}
\toprule
\textbf{Method} & \multicolumn{2}{c}{\textbf{Q1}} & \multicolumn{2}{c}{\textbf{Q2}} & \multicolumn{2}{c}{\textbf{Q3}} & \multicolumn{2}{c}{\textbf{Q4}} \\
\cmidrule(lr){2-3} \cmidrule(lr){4-5} \cmidrule(lr){6-7} \cmidrule(lr){8-9}
 & \scriptsize rL2 (\%) & \scriptsize NLL & \scriptsize rL2 (\%) & \scriptsize NLL & \scriptsize rL2 (\%) & \scriptsize NLL & \scriptsize rL2 (\%) & \scriptsize NLL \\
\midrule
\multicolumn{9}{l}{\emph{Elasticity (2D)}} \\
\midrule
Ensemble & $\mathbf{1.08}$ & $\mathbf{2.59}$ & $\mathbf{1.09}$ & $\mathbf{2.23}$ & $\mathbf{1.35}$ & $\mathbf{2.43}$ & $\mathbf{1.42}$ & $\mathbf{2.56}$ \\
MC Dropout & \valpmul{1.39}{0.11} & \valpm{19.04}{3.48} & \valpmul{1.42}{0.12} & \valpm{19.66}{3.32} & \valpmul{1.72}{0.13} & \valpm{20.05}{2.93} & \valpmul{1.81}{0.12} & \valpm{20.76}{3.22} \\
PNO & \valpm{2.20}{0.15} & \valpm{2.93}{0.09} & \valpm{2.23}{0.13} & \valpm{3.00}{0.07} & \valpm{2.51}{0.15} & \valpm{3.14}{0.08} & \valpm{2.62}{0.13} & \valpm{3.25}{0.06} \\
Perturbation & \valpm{1.99}{0.13} & \valpmul{2.71}{0.04} & \valpm{2.01}{0.14} & \valpm{2.75}{0.04} & \valpm{2.30}{0.14} & \valpmul{2.91}{0.07} & \valpm{2.40}{0.16} & \valpm{3.02}{0.07} \\
LUNO-LA & \valpm{1.49}{0.15} & \valpmbf{2.59}{0.28} & \valpm{1.52}{0.16} & \valpmul{2.60}{0.30} & \valpm{1.82}{0.16} & \valpm{2.93}{0.41} & \valpm{1.92}{0.18} & \valpm{3.09}{0.46} \\
DVE-spatial & \valpm{4.78}{0.19} & \valpm{9.69}{0.59} & \valpm{4.89}{0.20} & \valpm{9.68}{0.69} & \valpm{5.44}{0.27} & \valpm{10.71}{0.82} & \valpm{5.88}{0.27} & \valpm{11.88}{0.84} \\
\textbf{REEF-GP} & \valpm{1.49}{0.15} & \valpm{2.73}{0.09} & \valpm{1.52}{0.16} & \valpm{2.69}{0.08} & \valpm{1.82}{0.16} & \valpm{2.92}{0.11} & \valpm{1.92}{0.18} & \valpmul{2.97}{0.13} \\
\midrule
\multicolumn{9}{l}{\emph{Airfoil (2D)}} \\
\midrule
Ensemble & $\mathbf{0.85}$ & $\underline{-3.60}$ & $\mathbf{0.95}$ & $\mathbf{-3.56}$ & $\mathbf{1.13}$ & $\mathbf{-3.49}$ & $\mathbf{1.31}$ & $-3.19$ \\
MC Dropout & \valpm{1.38}{0.22} & \valpm{-3.49}{0.13} & \valpm{1.40}{0.19} & \valpmul{-3.45}{0.13} & \valpm{1.60}{0.18} & \valpm{-3.36}{0.14} & \valpm{1.94}{0.27} & \valpm{-3.22}{0.20} \\
PNO & \valpm{2.06}{1.40} & \valpm{-3.36}{0.65} & \valpm{2.12}{1.31} & \valpm{-3.36}{0.63} & \valpm{2.39}{1.47} & \valpm{-3.26}{0.68} & \valpm{2.73}{1.41} & \valpm{-3.13}{0.63} \\
Perturbation & \valpm{3.20}{0.19} & \valpm{-3.35}{0.05} & \valpm{3.13}{0.16} & \valpm{-3.36}{0.05} & \valpm{3.41}{0.19} & \valpm{-3.31}{0.05} & \valpm{3.60}{0.22} & \valpmul{-3.23}{0.06} \\
LUNO-LA & \valpm{1.03}{0.10} & \valpm{-3.37}{0.52} & \valpmul{1.13}{0.13} & \valpm{-3.28}{0.57} & \valpmul{1.30}{0.11} & \valpm{-2.92}{1.04} & \valpmul{1.51}{0.17} & \valpm{-2.68}{1.16} \\
DVE-spatial & \valpm{1.88}{0.27} & \valpm{17.36}{6.55} & \valpm{1.97}{0.26} & \valpm{18.01}{6.45} & \valpm{2.14}{0.25} & \valpm{20.48}{7.39} & \valpm{2.38}{0.20} & \valpm{24.14}{9.44} \\
\textbf{REEF-GP} & \valpmul{1.03}{0.10} & \valpmbf{-3.61}{0.05} & \valpmul{1.13}{0.13} & \valpmbf{-3.56}{0.07} & \valpmul{1.30}{0.11} & \valpmul{-3.47}{0.07} & \valpmul{1.51}{0.17} & \valpmbf{-3.41}{0.07} \\
\midrule
\multicolumn{9}{l}{\emph{Pipe (2D)}} \\
\midrule
Ensemble & $\mathbf{0.38}$ & $\mathbf{-6.27}$ & $\mathbf{0.31}$ & $\mathbf{-6.63}$ & $\mathbf{0.39}$ & $\mathbf{-6.58}$ & $\mathbf{0.60}$ & $\mathbf{-5.80}$ \\
MC Dropout & \valpm{0.54}{0.05} & \valpmul{-5.59}{0.06} & \valpm{0.52}{0.04} & \valpmul{-5.80}{0.04} & \valpm{0.68}{0.07} & \valpmul{-5.68}{0.11} & \valpm{0.84}{0.02} & \valpmul{-5.25}{0.16} \\
PNO & \valpm{0.65}{0.06} & \valpm{-5.20}{0.04} & \valpm{0.63}{0.05} & \valpm{-5.20}{0.04} & \valpm{0.77}{0.06} & \valpm{-5.15}{0.04} & \valpm{1.04}{0.07} & \valpm{-4.92}{0.07} \\
Perturbation & \valpm{2.06}{0.19} & \valpm{-4.33}{0.08} & \valpm{2.48}{0.24} & \valpm{-4.29}{0.09} & \valpm{2.48}{0.26} & \valpm{-4.22}{0.11} & \valpm{2.97}{0.30} & \valpm{-4.05}{0.11} \\
LUNO-LA & \valpmul{0.47}{0.05} & \valpm{-4.69}{0.15} & \valpmul{0.40}{0.03} & \valpm{-4.67}{0.16} & \valpmul{0.49}{0.06} & \valpm{-4.64}{0.16} & \valpmul{0.72}{0.03} & \valpm{-4.54}{0.14} \\
DVE-spatial & \valpm{3.34}{0.95} & \valpm{0.12}{1.04} & \valpm{3.73}{0.84} & \valpm{0.72}{1.77} & \valpm{3.55}{0.90} & \valpm{0.26}{1.54} & \valpm{4.13}{0.88} & \valpm{1.31}{2.16} \\
\textbf{REEF-GP} & \valpmul{0.47}{0.05} & \valpm{-4.72}{0.05} & \valpmul{0.40}{0.03} & \valpm{-4.85}{0.03} & \valpmul{0.49}{0.06} & \valpm{-4.79}{0.05} & \valpmul{0.72}{0.03} & \valpm{-4.57}{0.03} \\
\midrule
\multicolumn{9}{l}{\emph{ShapeNet Car (3D)}} \\
\midrule
Ensemble & $\mathbf{5.81}$ & $\underline{3.13}$ & $\mathbf{6.84}$ & $\underline{4.26}$ & $\mathbf{8.55}$ & $\underline{4.48}$ & $\mathbf{9.77}$ & $\underline{4.81}$ \\
MC Dropout & \valpm{7.19}{0.74} & \valpm{14.77}{4.73} & \valpm{8.02}{0.53} & \valpm{21.25}{5.54} & \valpm{9.84}{0.54} & \valpm{17.03}{3.67} & \valpm{11.15}{0.57} & \valpm{15.86}{3.09} \\
PNO & \valpm{12.50}{1.71} & \valpm{4.35}{1.24} & \valpm{12.08}{1.45} & \valpm{4.54}{1.21} & \valpm{14.61}{1.83} & \valpm{5.01}{1.47} & \valpm{15.89}{2.03} & \valpm{6.04}{2.19} \\
Perturbation & \valpm{6.65}{0.13} & \valpm{6.61}{0.65} & \valpm{7.62}{0.13} & \valpm{10.79}{1.14} & \valpm{9.48}{0.15} & \valpm{15.17}{1.70} & \valpm{10.79}{0.13} & \valpm{21.41}{1.77} \\
LUNO-LA & \valpm{6.59}{0.11} & \valpm{19.79}{13.47} & \valpm{7.64}{0.13} & \valpm{30.44}{21.51} & \valpm{9.51}{0.17} & \valpm{37.41}{25.97} & \valpm{10.79}{0.12} & \valpm{46.99}{33.15} \\
DVE-spatial & \valpm{10.67}{0.29} & \valpm{8.27}{1.90} & \valpm{11.12}{0.27} & \valpm{9.66}{2.65} & \valpm{14.17}{0.23} & \valpm{12.30}{3.51} & \valpm{16.46}{0.29} & \valpm{14.75}{4.15} \\
\textbf{REEF-GP} & \valpmul{6.47}{0.12} & \valpmbf{2.29}{0.02} & \valpmul{7.54}{0.14} & \valpmbf{2.68}{0.09} & \valpmul{9.42}{0.18} & \valpmbf{3.00}{0.11} & \valpmul{10.71}{0.14} & \valpmbf{3.42}{0.17} \\
\midrule
\multicolumn{9}{l}{\emph{Ahmed Body (3D)}} \\
\midrule
Ensemble & $\mathbf{5.16}$ & $\mathbf{3.90}$ & $\mathbf{6.35}$ & $\underline{4.77}$ & $\mathbf{5.83}$ & $\underline{4.03}$ & $\mathbf{6.60}$ & $\mathbf{4.09}$ \\
MC Dropout & \valpm{6.21}{0.46} & \valpm{10.75}{3.22} & \valpm{7.21}{0.46} & \valpm{18.75}{5.92} & \valpm{6.71}{0.39} & \valpm{13.13}{3.39} & \valpm{7.61}{0.43} & \valpm{17.75}{8.58} \\
PNO & \valpm{11.05}{1.05} & \valpm{5.86}{0.80} & \valpm{11.63}{0.95} & \valpm{12.70}{3.55} & \valpm{11.61}{0.78} & \valpm{5.13}{0.52} & \valpm{12.07}{1.10} & \valpm{9.84}{2.26} \\
Perturbation & \valpm{6.74}{0.26} & \valpm{5.42}{0.45} & \valpm{7.98}{0.13} & \valpm{7.97}{1.15} & \valpm{7.15}{0.18} & \valpm{6.47}{0.68} & \valpm{8.15}{0.17} & \valpm{7.66}{1.39} \\
LUNO-LA & \valpmul{5.74}{0.12} & \valpm{10.12}{5.77} & \valpm{6.94}{0.08} & \valpm{26.00}{17.84} & \valpmul{6.48}{0.16} & \valpm{9.45}{5.13} & \valpm{7.33}{0.23} & \valpm{22.22}{14.74} \\
DVE-spatial & \valpm{13.17}{1.52} & \valpm{8.02}{1.49} & \valpm{13.93}{1.68} & \valpm{8.93}{2.19} & \valpm{13.00}{1.48} & \valpm{6.14}{1.12} & \valpm{14.05}{1.69} & \valpm{8.50}{2.10} \\
\textbf{REEF-GP} & \valpm{5.82}{0.15} & \valpmul{4.06}{0.15} & \valpmul{6.92}{0.09} & \valpmbf{4.34}{0.08} & \valpm{6.49}{0.17} & \valpmbf{4.01}{0.16} & \valpmul{7.29}{0.23} & \valpmul{4.42}{0.06} \\
\bottomrule
\end{tabular}
}
\end{table}

\paragraph{OOD Evaluation Results.}
Table~\ref{tab:results_quantile_ood} reports the quantitative performance of all baselines partitioned across the four geometric MMD quartiles. A robust UQ method should seamlessly scale its predictive uncertainty as the test geometry deviates from the training manifold, resulting in a stable NLL even as the deterministic error (rL2) naturally increases. 

Across all benchmarks, REEF-GP demonstrates strong out-of-distribution stability. For instance, in the \emph{ShapeNet Car} dataset, as geometries shift from Q1 (most similar to the training samples) to Q4 (most different from training samples), REEF-GP's NLL increases only slightly and it remains the best-calibrated method in every quartile. In contrast, other post-hoc methods suffer calibration collapse under identical shifts: Perturbation's NLL and and LUNO-LA's NLL more than triple between Q1 and Q4, and similar trends are observed on the Ahmed Body benchmark. Furthermore, REEF-GP consistently delivers competitive metrics with deep ensembles.

\begin{figure}[h] 
    \centering 
        \includegraphics[width=\linewidth]{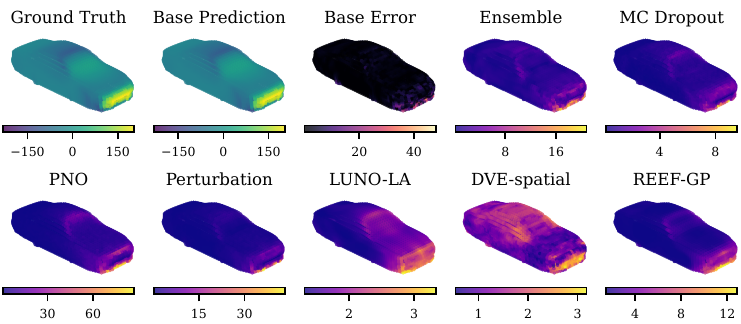} 
    \caption{Predictive standard deviation fields on a test \emph{Car} sample from Q1.} 
    \label{fig:car_q1} 
\end{figure}

\begin{figure}[b] 
    \centering 
        \includegraphics[width=\linewidth]{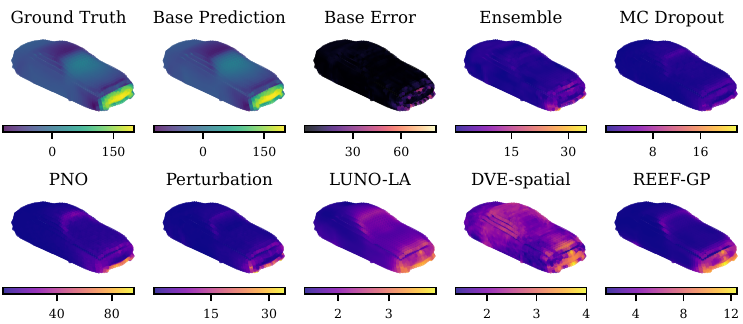} 
    \caption{Predictive standard deviation fields on a test \emph{Car} sample from Q2.}  
    \label{fig:car_q2} 
\end{figure}

\begin{figure}[b] 
    \centering 
        \includegraphics[width=\linewidth]{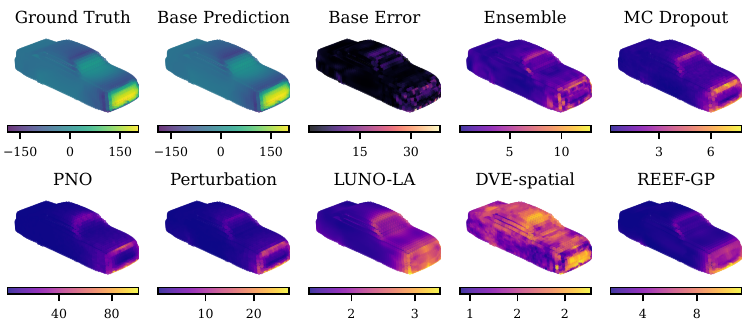} 
    \caption{Predictive standard deviation fields on a test \emph{Car} sample from Q3.} 
    \label{fig:car_q3} 
\end{figure}

\begin{figure}[b] 
    \centering 
        \includegraphics[width=\linewidth]{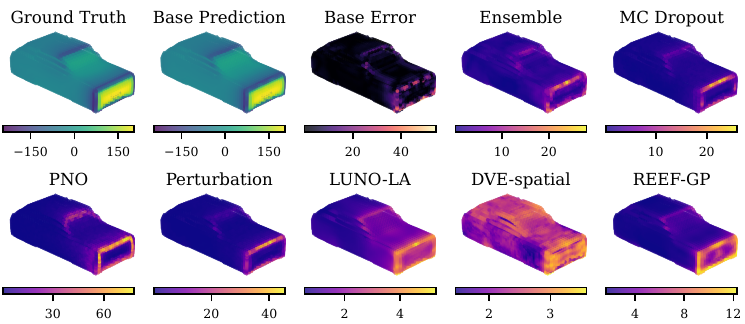} 
    \caption{Predictive standard deviation fields on a test \emph{Car} sample from Q4.}  
    \label{fig:car_q4} 
\end{figure}

\begin{figure}[b] 
    \centering 
        \includegraphics[width=\linewidth]{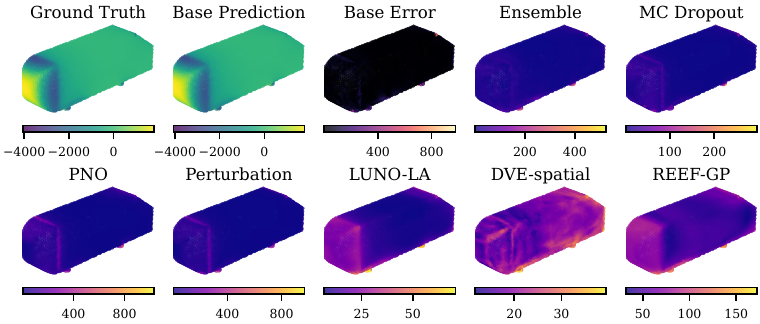} 
    \caption{Predictive standard deviation fields on a test \emph{Ahmed} sample from Q1.}  
    \label{fig:ahmed_q1} 
\end{figure}

\begin{figure}[b] 
    \centering 
        \includegraphics[width=\linewidth]{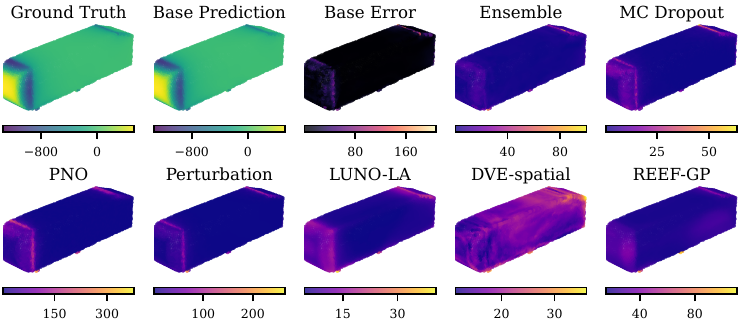} 
    \caption{Predictive standard deviation fields on a test \emph{Ahmed} sample from Q2.} 
    \label{fig:ahmed_q2} 
\end{figure}

\begin{figure}[b] 
    \centering 
        \includegraphics[width=\linewidth]{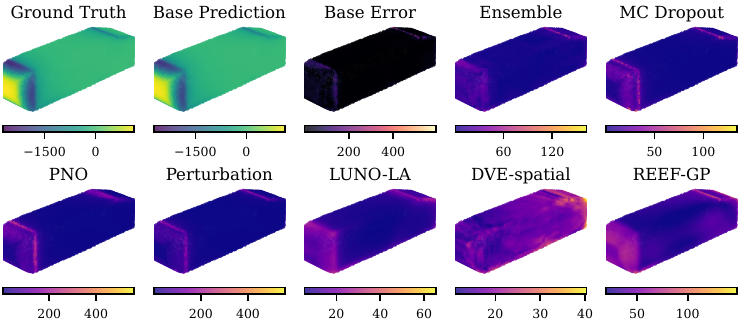} 
    \caption{Predictive standard deviation fields on a test \emph{Ahmed} sample from Q3.} 
    \label{fig:ahmed_q3} 
\end{figure}

\begin{figure}[b] 
    \centering 
        \includegraphics[width=\linewidth]{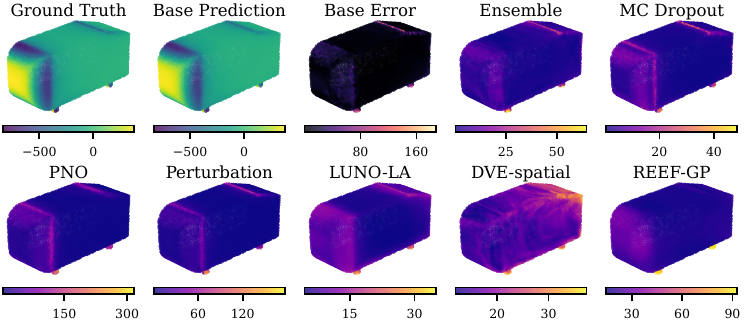} 
    \caption{Predictive standard deviation fields on a test \emph{Ahmed} sample from Q4.}  
    \label{fig:ahmed_q4} 
\end{figure}

\paragraph{Qualitative OOD Uncertainty.}
Figures~\ref{fig:car_q1} through \ref{fig:car_q4} and Figures~\ref{fig:ahmed_q1} through \ref{fig:ahmed_q4} visualize the spatial distribution of uncertainty for the 3D benchmarks as the test geometries become increasingly distinct from the training set (progressing towards Q4). REEF-GP's predicted uncertainty remains coherent with the base model error regions over the geometry. 

\clearpage


\end{document}